\UseRawInputEncoding
\documentclass[journal]{IEEEtran}
\usepackage[T1]{fontenc}
\usepackage{graphicx}
\usepackage{times}
\usepackage{helvet}
\usepackage{courier}
\usepackage{amsmath}
\usepackage{algorithm}
\usepackage{algorithmic}
\usepackage{csquotes} 
\usepackage{color}
\usepackage{paralist}
\usepackage{amssymb}
\usepackage{indentfirst}
\usepackage{subfigure}
\usepackage{float}
\usepackage{multirow}
\usepackage{cite}
\usepackage{mathrsfs}

\usepackage{mathrsfs}
\usepackage[colorlinks, linkcolor=red,  anchorcolor=blue, citecolor=blue]{hyperref}
\usepackage{textcomp,booktabs}
\usepackage{amssymb}
\usepackage{pifont}
\newcommand{\cmark}{\ding{51}}%

\usepackage{makecell}

\usepackage{colortbl}
\definecolor{mygray}{gray}{.9}
\definecolor{mypink}{rgb}{.99,.91,.95}
\definecolor{mycyan}{cmyk}{.3,0,0,0}

\begin{document}

\title{Beyond Greedy Search: Tracking by Multi-Agent Reinforcement Learning-based Beam Search}  

\author{Xiao Wang, \emph{Member, IEEE}, Zhe Chen, \emph{Member, IEEE}, Bo Jiang, Jin Tang, Bin Luo, \emph{Senior Member, IEEE}, \\ and Dacheng Tao, \emph{Fellow, IEEE}    
\thanks{Xiao Wang, Bo Jiang, Jin Tang and Bin Luo are with the School of Computer Science and Technology, Anhui University, Hefei 230601, China. Tang is also with Cognitive Computing Research Center of Anhui University. } 
\thanks{Zhe Chen is with The University of Sydney, Australia. } 
\thanks{Dacheng Tao is with the JD Explore Academy, China and also with the University of Sydney, Australia. } 
\thanks{Corresponding author: Jin Tang and Zhe Chen.} 
\thanks{Email: \{xiaowang, jiangbo, tangjin, luobin\}@ahu.edu.cn, \{zhe.chen1, dacheng.tao\}@sydney.edu.au.}}

\markboth{IEEE Transactions on Image Processing} 
{Shell \MakeLowercase{\textit{et al.}}: Bare Demo of IEEEtran.cls for IEEE Journals}

\maketitle

\begin{abstract}
To track the target in a video, current visual trackers usually adopt greedy search for target object localization in each frame, that is, the candidate region with the maximum response score will be selected as the tracking result of each frame. However, we found that this may be not an optimal choice, especially when encountering challenging tracking scenarios such as heavy occlusion and fast motion. In particular, if a tracker drifts, errors will be accumulated and would further make response scores estimated by the tracker unreliable in future frames. To address this issue, we propose to maintain multiple tracking trajectories and apply beam search strategy for visual tracking, so that the trajectory with fewer accumulated errors can be identified. Accordingly, this paper introduces a novel multi-agent reinforcement learning based beam search tracking strategy, termed BeamTracking. It is mainly inspired by the image captioning task, which takes an image as input and generates diverse descriptions using beam search algorithm. Accordingly, we formulate the tracking as a sample selection problem fulfilled by multiple parallel decision-making processes, each of which aims at picking out one sample as their tracking result in each frame. Each maintained trajectory is associated with an agent to perform the decision-making and determine what actions should be taken to update related information. More specifically, using the classification-based tracker as the baseline, we first adopt bi-GRU to encode the target feature, proposal feature, and its response score into a unified state representation. The state feature and greedy search result are then fed into the first agent for independent action selection. Afterwards, the output action and state features are fed into the subsequent agent for diverse results prediction. When all the frames are processed, we select the trajectory with the maximum accumulated score as the tracking result. Extensive experiments on seven popular tracking benchmark datasets validated the effectiveness of the proposed algorithm.
\end{abstract}


\IEEEpeerreviewmaketitle

\section{Introduction}

\begin{figure*}
\center
\includegraphics[width=7in]{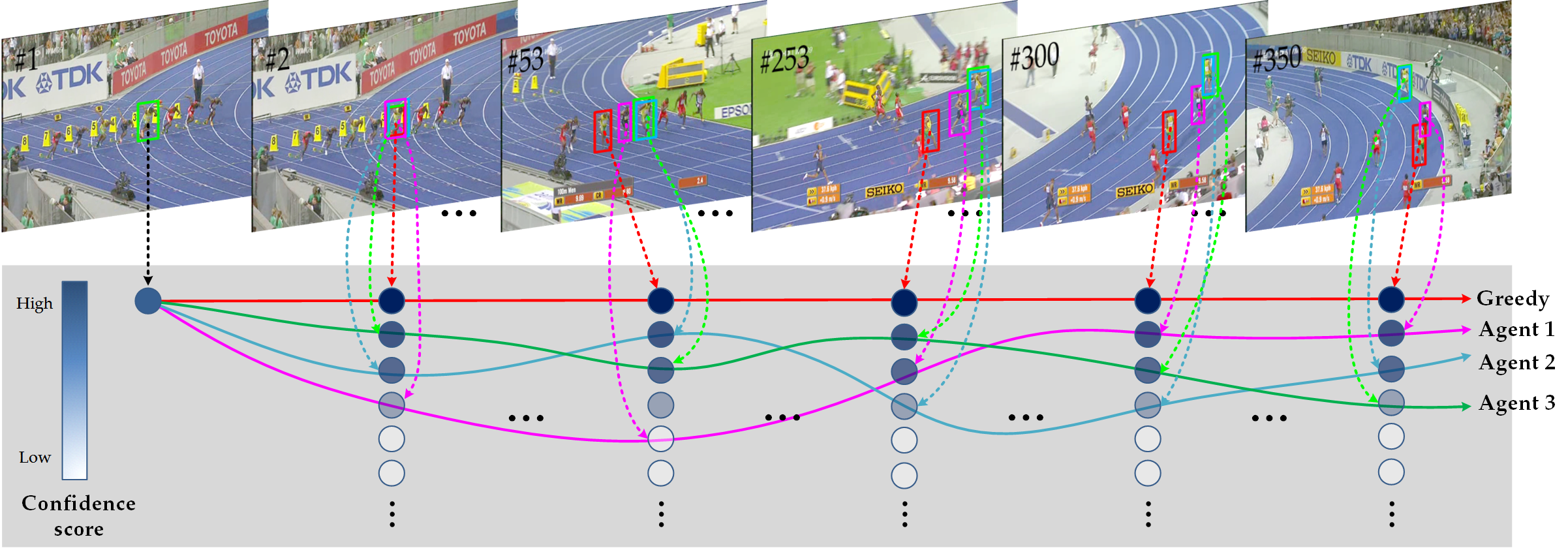}
\caption{Comparison between greedy search and our proposed MARL based beam search for visual tracking. Previous trackers usually select the location with maximum response score as their result in each frame, and predict the trajectory of target object in a greedy manner (like the red trajectory). In contrast, our proposed MARL based beam search  algorithm treats the visual tracking task as a multi-trajectory selection problem, which contains the local-global-proposal generation module, state generation/update module, and beam search policy networks. It selects and maintains the locations with reasonable scores instead of the maximum one for each frame, and chooses the trajectory with the highest score after summation. }
\label{motivation}
\end{figure*}

\IEEEPARstart{T}{he} goal of visual tracking is to search for target object in subsequent frames according to its initial state in the first frame. With great progress achieved in recent years, tracking techniques are widely used in many applications, such as video surveillance, robots and autopilot. Nevertheless, due to various challenging factors, such as heavy occlusion and fast motion, the performance of existing trackers in a complex environment is still far from satisfactory.

Existing trackers generally adopt either online or offline learning schemes to conduct tracking~\cite{lu2018visualSurvey, li2018deepTrackSurvey, marvasti2019deepSurvey, zhang2020long, guo2021exploring, wang2021visevent, hong2015multi}. Online learning trackers update their appearance models on-the-fly, while offline learning trackers use the appearance models pre-trained on other datasets. However, according to our observation, both existing online learning and offline learning trackers apply a greedy search strategy to perform tracking, that is, only the result with the highest confidence score predicted by the tracker at each frame will be considered as the final tracking result. 
Although this greedy search strategy can achieve good performance on simple videos, existing methods can be still very vulnerable to challenging factors like occlusions and motion blurs.
For example, to update tracking models, online learning trackers can only use the sample initialized in the first frame and noisy results obtained during tracking, making them difficult to identify and correct unreliable target appearances learned from inaccurate results. For the offline trained trackers, their parameters are fixed and thus can not depict the variations of the target object comprehensively. 
Therefore, existing methods that force the tracker to locate the target and learn the target appearance only based on the maximum response scores can be still prone to drifting. This inspired us to think \emph{do we have to make the tracker predict the only one location for each frame in such a greedy manner?}

\begin{figure*}
\center
\includegraphics[width=7in]{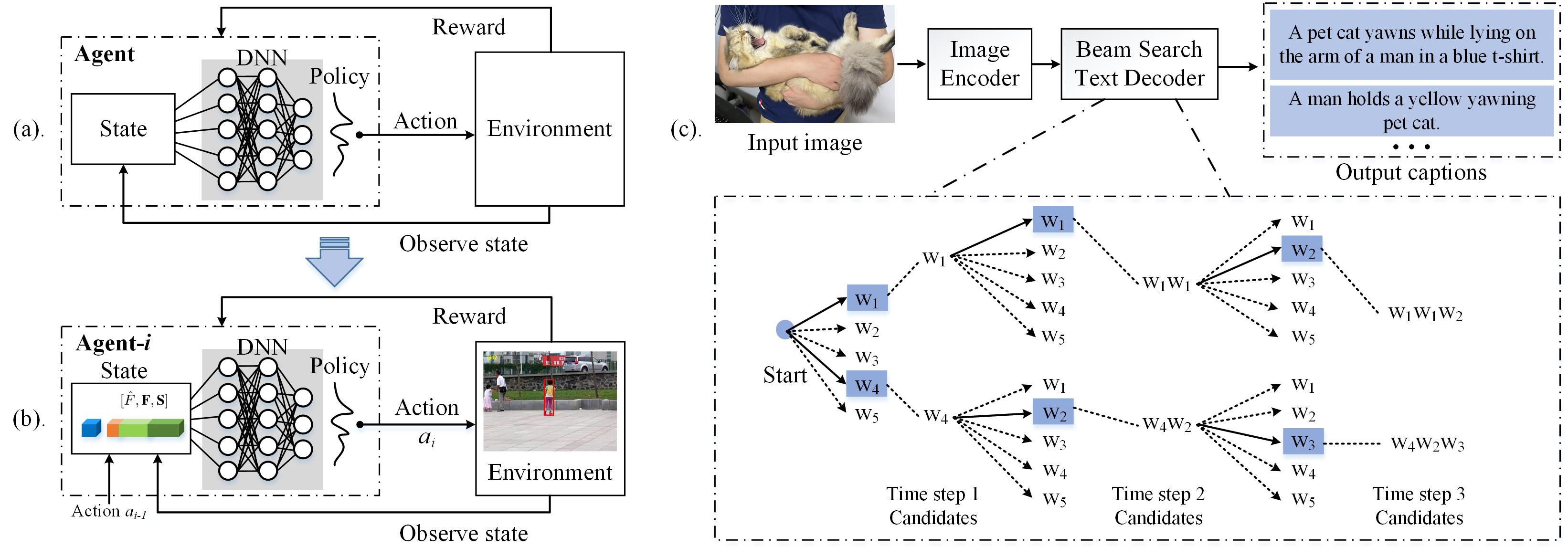}
\caption{
(a). The schematic of interaction between agent and the environment in the standard deep reinforcement learning. 
(b). The agent-environment interaction process in our case. 
(c). Illustration of beam search-based image captioning. Given the input image, an image encoder is usually adopted for feature extraction, then, a language decoder is used for captioning generation. Note that, the beam search algorithm is often utilized to obtain diverse language descriptions, which inspired us to design the MARL-based beam search for visual object tracking.}
\label{captionBeamSearch}
\end{figure*}

Recently, image captioning~\cite{Aneja_2018_CVPR, hossain2019captionSurvey}, which takes an image as input and outputs a sentence to describe the contents of a given image, draws increasing attention among researchers. More importantly, some researchers adopt the beam search algorithm \footnote{\url{https://en.wikipedia.org/wiki/Beam_search}} to maintain multiple candidate words for each time step and finally obtain diverse captions, as illustrated in Fig. \ref{captionBeamSearch}. Inspired by this procedure, for the visual object tracking task, we can relax the greedy search (select one maximum response region) of target object in each frame by selecting multiple reasonable response regions. Although directly introducing the beam search into visual tracking is an intuitive approach, however, the vanilla beam search algorithm which selects the top-\emph{k} candidates could still deliver similar effects with greedy algorithms. 
The top-\emph{k} candidate samples in practical tracking can be very similar to each other. Therefore, the original beam search mechanism would not be superior in tracking compared with the regular greedy search algorithms. As a result, how to design a more suitable beam search algorithm for visual tracking remains an open question.

In this paper, we introduce a novel multi-agent reinforcement learning (MARL) based beam search policy to magnify the benefits of beam search for tracking and alternate the commonly used greedy search-based strategy. The comparison between greedy search and the proposed MARL-based beam search is illustrated in Fig. \ref{motivation}. Specifically, we formulate visual tracking as a sample selection problem. Different from the trackers~\cite{hare2015struck, Nam2015Learning, Jung_2018_ECCV, wangmfgnet2022, zhang2021learn} that also perform tracking based on sample selection, we introduce a multi-agent reinforcement learning framework for tracking the target. In the proposed framework, we employ different agents and apply a multi-agent decision making process to estimate the state of the target object based on candidate states in each frame. We define each candidate state according to a proposal (a bounding box area that potentially contains the target) and extracts its state representation based on the confidence score of the proposal, the CNN features within the proposal, and the CNN features of the target. Proposals and their confidences are obtained from a joint local and global search architecture according to previous tracking locations and target-aware attention regions~\cite{wang2019GANTrack}. A recurrent neural network (bi-directional GRU~\cite{chung2014GRU}) is further utilized to comprehensively encode candidate states into a unified state feature. Then, we introduce parallel decision making processes which assign different policy networks for various agents. A policy network takes the unified state feature as input and selects the best candidate state as the tracking result for the agent in current frame. To boost the communication between nearby agents, we also feed the actions chosen by the previous agent into the subsequent one. After selecting different target states with different agents, we can then apply the beam search to update the maintained multiple tracking trajectories until current frame. These operations are executed until the end of a video sequence. Lastly, the best-scored trajectory will be chosen as the tracking result of current video. Following the reinforcement learning paradigm, we optimize the proposed multi-agent beam search network with the policy gradient method (Proximal Policy Optimization algorithm, PPO~\cite{schulman2017proximal}). The overall pipeline of our tracker can be found in Fig. \ref{pipeline}.

Compared with existing visual trackers, the features and differences of this work can be concluded as follows: 
\textbf{{(1). Frame-level vs Trajectory-level decision}:}
Different from previous algorithms which employ a frame-level decision strategy for visual tracking, we adopt the MARL-based beam search to exploit the trajectory-level decision. In practice, our tracker can realize global reasoning for some challenging frames. 
\textbf{{(2). Ensemble of multi-trackers vs Beam search-based single tracker}:} 
Traditional ensemble learning-based trackers usually adopt multiple trackers which may bring extremely high time and space complexity. In contrast, our proposed beam search can be integrated with only one tracker but also achieves high-performance tracking. 
\textbf{{(3). Single-agent vs Multi-agent RL}:}  
Previous RL based tracking algorithms all follow a single-agent setting, while our tracker is developed based on a multi-agent RL paradigm. By cooperating multiple agents in our method, the tracking performance can be further improved. 

To sum up, the contributions of this paper can be summarized in the following three aspects: \footnote{A demo video for this work is available at: \url{https://youtu.be/f1yiYv-SJyY}}  

$\bullet$ We analyze the limitations of greedy search used in regular visual tracking framework and propose to conduct tracking with multi-agent reinforcement learning based beam search strategy, termed BeamTracking. To the best of our knowledge, this is the first attempt to perform beam search based single object tracking task under a multi-agent reinforcement learning framework.

$\bullet$ We formulate the visual tracking task as a sample selection problem that can be tackled with multiple parallel Markov decision making processes. We propose a multi-agent reinforcement learning framework to fulfill the sequential decision-making problem. 

$\bullet$ We integrate the MARL beam search strategy into multiple trackers and conduct experiments on multiple popular tracking benchmark datasets. These results fully validated the effectiveness and generalization of our proposed approach.

\section{Related Work} 

In this section, we will give a review on greedy search based, reinforcement learning based, and ensemble learning based trackers. Due to the limited space in this paper, the following survey papers~\cite{marvasti2019deepSurvey, li2018deepTrackSurvey, lu2018visualSurvey, li2013survey, Smeulders2014Visual, chen2015KCFsurvey} and paper list \footnote{\url{https://github.com/wangxiao5791509/Single_Object_Tracking_Paper_List}} are recommended to find more related trackers.  

\textbf{Greedy Search based Tracking.~~~} 
The classification based trackers take the visual tracking as a binary classification problem, and they learn a classifier online and discriminate whether the given proposal is a target object or background. Some traditional trackers like Struck~\cite{hare2015struck}, SOWP~\cite{Kim2016SOWP} and deep learning trackers MDNet~\cite{Nam2015Learning} are very popular in the tracking community. Pu et al.~\cite{pu2020LRMANet} propose a recurrent memory activation network (RMAN) to exploit the untapped temporal coherence of the target appearance for visual tracking. Some works attempt to improve the overall performance from different perspectives, such as ensemble learning~\cite{han2017branchout}, improve tracking efficiency~\cite{Jung_2018_ECCV}, meta-learning~\cite{Park_2018_ECCV} and joint local and global search~\cite{wang2019GANTrack}.

Benefiting from the offline training with large-scale datasets, the Siamese network based trackers achieve good performance on recent tracking benchmarks. SiamFC~\cite{Bertinetto2016SiameseFC} and SINT~\cite{Tao2016Siamese} are early works to introduce the Siamese network by designing fully convolutional architecture and instance matching for tracking, respectively. Later, the modules proposed in object detection community are modified for tracking task like RPN module~\cite{li2018siamRPN, li2018siamrpn++}, re-detection module~\cite{2020siamRCNN, yan2019skimming, qi2020siamese, wang2021tnl2k}, anchor-free module~\cite{zhangocean, xu2020siamfc++}, adversarial training~\cite{Wang_2018_CVPR, SongYiBing_2018_CVPR}, etc. Besides, there are also many works introduce new learning schemes, for example, meta-learning~\cite{Park_2018_ECCV, wang2020metatrack}, reinforcement learning~\cite{Yun2017Action, ren2018deepDRL, luo2018end}, unsupervised learning~\cite{wang2019unsupervisedTrack}.

Many researchers devote themselves to improving the accuracy and efficiency of visual tracking through more powerful feature representation learning, knowledge distill, hierarchical attention, Transformer, etc. Specifically, Liang et al.~\cite{liang2019localSiamFastTrack} propose the Local Semantic Siamese (LSSiam) network to extract more fine-grained and partial information for visual tracking. A tracking-specific distillation strategy is considered by Shen et al.~\cite{shen2021DSNTrack} to achieve small, fast and accurate student tracker learning from large Siamese trackers. Lu et al.~\cite{lu2020deepShrinkageLoss} propose a shrinkage loss to address the extremely imbalanced pixel-to-pixel differences in tracking, therefore, the enhanced trackers can better distinguish target objects from the background. Shen et al.~\cite{shen2017fastDetRefinement} propose a  minimum output sum of squared error filter which can adaptively refine the tracking targets.  A hierarchical attention mechanism is proposed in~\cite{shen2019HAttSiamTrack} which can adaptively fuse multi-scale response maps for Siamese visual tracking. The potential connections among the training instances are exploited by the quadruplet deep network proposed in~\cite{dong2019quadrupletFastTrack}, which achieves a more powerful representation. Recent works demonstrate that the new architectures like Transformers~\cite{vaswani2017transformer} are also improve the Siamese tracker significantly~\cite{chen2021transTrack,  wang2021transTrack}. Although these trackers achieve good performance on some benchmarks, however, nearly all of these trackers adopt a greedy search strategy for final tracking (including the Correlation Filter based trackers~\cite{zhang2020sparse, han2020fast, zheng2020multi}), which will make it difficult for the tracker to revise its past predictions based on future observations. In this paper, we propose a novel MARL-based beam search algorithm to achieve multiple-trajectory tracking, which demonstrates significant improvement compared with regular greedy search.

\textbf{Reinforcement Learning based Tracking.~~~} 
Deep reinforcement learning (DRL) draws more and more attention to computer vision researchers. There are already some works proposed to introduce the DRL into the visual tracking community~\cite{Yun2017Action, zhang2017deep, luo2018end, supancic2017tracking, ren2018collaborative, dong2018hyperparameter, luo2016model, Wang_2018_CVPR, chen2018realACT}. Specifically, Yun et al.~\cite{Yun2017Action} propose a tracker that is controlled by sequentially pursuing actions learned by deep reinforcement learning. The use of reinforcement learning enables even partially labeled data to be successfully utilized for semi-supervised learning. Chen et al.~\cite{chen2018realACT} take tracking as a continuous reinforcement learning problem and train an actor-critic network to move the bounding box directly to locate the target object.  Zhang et al.~\cite{zhang2017deep} also treat the tracking problem as a sequential decision-making process and historical semantics encode highly relevant information for future decisions. They formulate their model as a recurrent convolutional neural network agent that interacts with video over time and achieves good performance on tracking benchmark. James et al.~\cite{supancic2017tracking} propose to learn an optimal decision-making policy by formulating tracking as a partially observable decision-making process. Hence, their agent can adaptively decide where to look, when to reinitialize, and when to update its appearance model for the tracked object. Wang et al.~\cite{Wang_2018_CVPR} utilize RL to learn to generate hard instance samples and integrate them with Siamese tracker for robust tracking. Different from these trackers which utilize a single agent to optimize their target and adopt greedy search for visual tracking, in this work, we formulate the visual tracking as a multi-agent decision-making process and propose a novel beam search strategy to handle the issues caused by greedy search.

In contrast, multi-agent reinforcement learning attempts to achieve maximum reward through the communication of multiple agents. More interestingly, these agents may have different outcomes depending on what the other agents are doing~\cite{neto2005single}. MARL have been widely used in many computer vision tasks, such as multi-object tracking~\cite{rosello2018multi, ren2018collaborative, jiang2019multi}, action recognition~\cite{wu2019multiMARL} and traffic light control~\cite{bakker2010traffic, kuyer2008multiagent}. To the best of our knowledge, this paper is the first work to introduce the idea of multi-agent reinforcement learning for visual tracking.

\textbf{Ensemble Learning based Tracking.~~~} 
Seldom existing trackers model visual tracking from the view of multi-trajectory analysis~\cite{lee2015multihypothesis}~\cite{kim2015MHT}. Specifically, MTA is proposed in~\cite{lee2015multihypothesis} to conduct tracking by trajectory selection in the tracking procedure based on STRUCK~\cite{hare2016struck}. Their tracking performance is limited by manually designed features and untrainable multi-trajectory analysis. Wang et al. proposed the DeepMTA~\cite{wang2021deepmta} by introducing multi-trajectory analysis based on dynamic target-aware attention and trajectory evaluation networks. MHT~\cite{kim2015MHT} is developed based on empirically defined track tree construction and updating scheme. It also relies on hand-crafted features and detection results to solve multi-object tracking problems.

Some researchers resort to ensemble learning for high-performance tracking by fusing multiple experts. Specifically, the MEEM~\cite{zhang2014meem} is developed to address the model drift issue with a multi-expert restoration scheme. CF2~\cite{ma2015hierarchical} adaptively learns correlation filters on each convolutional layer to encode the target appearance and hierarchically infer the maximum response of each layer to locate targets. HDT~\cite{qi2016HDT} use different CNN features and hedge several CNN trackers into a stronger one with an adaptive Hedge method. Besides, the AdaBoost is used for tracking with ensemble learning, such as~\cite{avidan2007emtracking, bai2013randomizedET, grabner2006RTonlineBoost}. MCCT~\cite{wang2018multicueTrack} is proposed by Wang et al. for accurate object tracking, called the Multi-Cue Correlation filter based Tracker.~\cite{biresaw2014trackerFusion} fuses multiple trackers based on appropriately mixing the prior state of the trackers. It also updates the target appearance model according to the track-quality level.~\cite{khalid2016multiTracker} first group trackers into clusters based on the spatio-temporal pair-wise correlation, then the reverse-time analysis is employed to select the high-quality clutters for late fusion.~\cite{leang2018onlineSOT} presents a generic framework for combining or selecting online the different components of the processing chain of a set of trackers. HMMTxD is proposed in~\cite{vojir2016online} which fuses observations from complementary out-of-the box trackers and a detector with a hidden Markov model. Wang et al.~\cite{wangnaiyan2014ensemble} jointly learn the unknown trajectory of the target and the reliability of each tracker for ensemble-based tracking, termed FHMM. Xie et al.~\cite{xie2019multiFusion} achieve multi-tracker fusion by removing the bad trajectories based on the pair-wise correlation, which can be obtained between different tracker pairs. Different from these works, we propose a novel multi-agent reinforcement learning based beam search strategy for visual tracking, which can maintain multiple tracking results for each frame. Our tracker can realize video-level decisions which will be more beneficial for tracking.

\section{Preliminaries of DRL and Beam Search}
In this section, we give an introduction to the background knowledge of this work, including deep reinforcement learning and beam search.

\subsection{Deep Reinforcement Learning}  
As shown in Fig. \ref{captionBeamSearch} (a), the standard reinforcement learning framework contains two mainly components, i.e., the agent and the environment.
This framework encourages the agent to learn from returns (both reward and punishment) obtained by interacting with the environment. Specifically, the agent selects an action and the state of the environment will be changed with a certain probability (this probability is usually assumed to be 1 to simplify the scenario). A positive reward is given if the state is moving in the direction one would expect, and a negative reward otherwise. By continuously accumulating experience (i.e., the training data), the agent can continue to evolve to achieve a certain degree of intelligence.

Correspondingly, as shown in Fig. \ref{captionBeamSearch} (b), in our case, the agent is the policy network, the environment is the tracking status.
To achieve beam search-based tracking, we formulate visual tracking as a sample selection problem that can be tackled with Markov decision process (MDP). To describe an MDP, we introduce its basic elements: $(S, A, T, R, \gamma)$, where $S$ is a set of states $s_t$, $t$ is used to denote time step, $A$ is a set of actions $a_t$, $T$ is the transition probability of the next state given current state and action \footnote{In our case, given the current state and action, the transition probability is determined.}, $R$ is the reward function $r(s_t, a_t)$ for each intermediate time step $t$, and $\gamma$ is the discount factor which usually set as 0.9. The actions are taken from a probability distribution called policy $\pi$ given current state $(a_t \thicksim \pi (s_t))$. Under MDP, it is straightforward to apply the reinforcement learning technique to learn the policy $\pi$ or the tracking model. By maximizing the rewards, reinforcement learning can achieve better tracking performance. Many visual trackers are developed based on this idea~\cite{Yun2017Action, zhang2017deep, luo2018end, supancic2017tracking, ren2018collaborative, dong2018hyperparameter, luo2016model, luo2018end, Wang_2018_CVPR, chen2018realACT}. Different from previous works, this paper develops a novel beam search-based policy with multi-agent reinforcement learning for visual tracking, which can track the target object by analyzing multiple tracking results.

\subsection{Beam Search} 
As a greedy heuristic search algorithm, the Beam Search (BS) scheme~\cite{vijayakumar2016diverseBS, sun2017bidiBS} is prevalent in image captioning community. The BS adopts the breadth-first search to traverse the  search tree, that is to say, the top-$B$ highest-scoring partial hypotheses (also called \emph{beams}) are stores at each time step. Here, the $B$ is also termed the \emph{beam width}. Formally, at time step $t$, we denote the partial hypothesis (beam) as $Y_{[1:t] = (y_1, ... , y_t)}$, and denote a collection of $B$ beams is denoted as $Y_{[1:B][1:t]} = (Y_{1,[1:t]}, Y_{2,[1:t]}, ... , Y_{B,[1:t]})$. As shown in Fig. \ref{captionBeamSearch}, at the start step, the BS scheme is initialized with empty beams, $Y_{b,0} = (y_{b,0})$, where $y_{b,0} = \emptyset$. For each time step $t$, all possible beam extension $\mathcal{Y}_t = Y_{[1:B],[1:t-1]} \times \mathcal{Y}$ are considered and the top-$B$ high-scoring $t$-length beams among this expanded hypothesis space. The search for optimal updated beams $Y_{[1:B][1:t]}$ can be written as follows: 
\begin{equation}
\label{BSObjective} 
\underset{\mathcal{Y}_{t}}{\text { top-B }} \log \mathrm{P}\left(Y_{[1: t]}\right)=\sum_{i=1}^{t} \log \mathrm{P}\left(\mathbf{y}_{i} \mid \mathbf{y}_{i-1}, \ldots, \mathbf{y}_{1}\right). 
\end{equation}
It is worthy to note that each log probability term in Eq. (\ref{BSObjective}) can be obtained via the forward pass in the decoder network (such as the LSTM). Actually, the top-$B$ operation can be simply obtained by sorting the $B\mathcal{Y}_t$ values. A simple example is illustrated in Fig. \ref{captionBeamSearch}.

\section{Tracking by MARL based Beam Search  }

In this section, we first give an overview of our proposed visual tracking system. Then, we dive into the details of how we implement MARL based beam search for tracking. After that, we talk about the objective functions used in our tracking framework. Finally, we introduce the details in the tracking phase.

\subsection{Overview} 
As shown in Fig. \ref{pipeline}, we first collect candidate states of the target object for each frame based on binary classification based tracking framework. Specifically, the local and global proposals are extracted around previous tracking results and attention regions. The response scores of these proposals can be attained with an online learned classifier. Then, we concatenate the target object features, the features within proposals, and the response scores of proposals to form candidate states. Subsequently, a bi-directional GRU network is utilized to encode the states into one feature representation. In this procedure, the historical proposals extracted from previous frames are memorized to facilitate current prediction. For each frame, we employ policy networks, i.e., multiple agents, to output diverse actions for beam search-based tracking. The selections of different agents are enhanced by the communications between nearby agents. In another word, we input the action chosen by the previous agent into the subsequent one to boost the information propagation between different agents when selecting their individual action. As a result, this multi-agent system generates multi-trajectory tracking results. After spanning the entire video sequence, the response score of an obtained trajectory can be computed by simply summarizing the scores of proposals that belong to the corresponding trajectory. We select the trajectory with the maximum accumulated response score as the final tracking result of testing video sequence.

\subsection{Candidate States Generation via Local-Global-Search} 
To effectively apply beam search for tracking, we need multiple reliable but diverse tracking results. We achieve this goal by employing multiple agents and letting each agent find the best state of the target from multiple candidate states. We generate candidate states based on joint local and global search results. The local search provides accurate tracking results for simple videos but can be extremely vulnerable to challenging factors, such as fast motion, heavy occlusion, and out-of-view, which are frequently occurred in a practical tracking environment. The global search then complements the local search to improve the tracking robustness in these challenging scenarios.

For local search-based tracking, we generate candidate states based on the binary classification-based multi-domain convolutional neural network, i.e., MDNet~\cite{Nam2015Learning} and RT-MDNet~\cite{Jung_2018_ECCV}. In these trackers, dozens of proposals are sampled around previous tracking results. Then, these trackers feed sampled proposals into a binary classifier to select the best-scored proposal as the result of the current frame. After tracking, they generate positive and negative training samples and train their network on-the-fly. Different from MDNet, the RT-MDNet greatly improves the tracking efficiency by introducing a fast adaptive RoI-Align scheme for feature extraction of sampled proposals. In practice, by applying these trackers to perform a local search for the target, we can obtain a rich collection of proposals and their confidence scores, thus forming multiple candidate states.

\begin{figure*}[!thb]
\center
\includegraphics[width=7.1in]{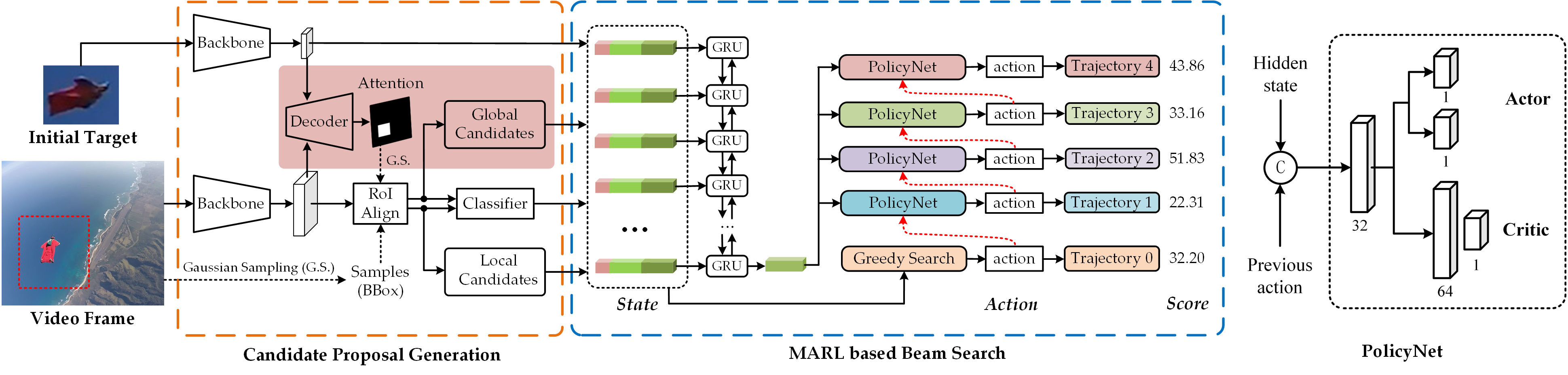}
\caption{Pipeline of the proposed MARL based beam search for visual tracking (The RT-MDNet~\cite{Jung_2018_ECCV} is adopted as an example). Specifically, it can be divided into two main modules, i.e., the candidate proposal generation (CPG) and MARL based beam search module (MARL-BS). The CPG is designed to output both local and global candidate search features, according to previous tracking results and target-aware attention network (TANet). The MARL-BS module encodes the features and scores of each proposal into a unified representation using a bi-directional GRU network, then feeds the output into multiple policy networks for diverse trajectory prediction. Note that, the inputs of each policy network are the state feature, and the action predicted by its previous policy network, therefore, they can output diverse locations.}
\label{pipeline}
\end{figure*}

For global search-based tracking, we mainly follow~\cite{wang2019GANTrack} and adopt Target-aware Attention Network (termed TANet) to locate the target object from the global view in this paper. In particular, we adopt ResNet-18~\cite{he2016deep}, which is pre-trained on ImageNet~\cite{russakovsky2015imagenet}, to extract features of initial target object and global video frame. The feature maps of target object are then utilized as the convolutional filter and implement convolutional operation on the feature maps of the global video frame. The processed feature maps will be fed into a decoder network gradually by skip-connections. The decoder network will generate the corresponding attention map where the target object related regions will be highlighted. With the generated attention of the target, we can sample global search-based proposals for the target. These proposals are also fed into the local search tracker to obtain confidence scores. This procedure is illustrated in Fig. \ref{pipeline}.

As mentioned in previous sections, the commonly used greedy search strategy that selects the proposal with the maximum response score as the result of the current frame can already work well on simple videos. However, this mechanism still suffers from drifting in challenging scenarios under the tracking-by-detection framework. To address this issue, we propose to perform a beam search on multi-trajectory information for more robust tracking. The multi-trajectory is obtained with a multi-agent reinforcement learning (MARL) framework, which will be introduced in the following sections.

\subsection{MARL Framework}
The introduction of MARL framework is to magnify the benefits of beam search for tracking. In general, the MARL employs different agents to perform tracking. Each agent follows a typical reinforcement learning framework. Therefore, we need to define the state representation, the action space, the decision making policy, and the reward function for training each agent. In this section, we will describe in detail how we define these concepts to fulfill tracking.

\subsubsection{\textbf{State Representation}}
As described in the introduction, we obtain the state representation for each agent based on a collection of candidate state representations. Candidate state representations are generated according to proposals by the joint local and global search. Confidence scores and CNN features define candidate state representations.

Formally, we introduce the following symbols to help describe state representations. Given a video frame, we first collect $n$ proposals $\textbf{P} = \{ p_1, p_2, ... , p_n \}$. Each $p_i$ is a 4-D vector describing the coordinates of the four corners of a bounding box area. For the collected proposals, we use the symbols $\textbf{S} = \{ s_1, s_2, ... , s_n \}$ to represent corresponding confidence scores predicted by the baseline visual tracker. Then, we denote $\textbf{F} = \{ F_1, F_2, ... , F_n \}$ as the features extracted from CNN for all the collected proposals. Each $F_i$ is originally a tensor with the size $C\times H \times W$, and we then reshape it into a 1-D vector. We also extract the visual feature (denoted as $\hat{F}$) of the target object initialized in the first frame. Many previous works have validated the importance of an initial target object patch for practical tracking~\cite{Tao2016Siamese, li2018siamRPN, zhipeng2019deeper, wang2019GANTrack}. Then, for the $i$-th proposal, we concatenate its confidence score $s_i$, the convolutional feature within its bounding box area $F_i$, and the initial target feature $\hat{F}$ into a unified feature vector. We use $[\hat{F}, \textbf{F}, \textbf{S}]$ to represent the collection of this concatenation operation according to the collection of proposals $\textbf{P}$.

After collecting the candidate state representations, we then extract a unified state representation for agents. In our MARL framework, each agent will learn a decision making policy to select a candidate state that best describes the target state in the current frame. To learn a good decision making policy, it is important that each agent can observe more about the environment that surrounds the target. Rather than modeling the whole image that introduces excessive computational costs, we propose that the candidate representations already provide informative but diverse observations about the surrounding environments, since proposals are generally close to the target. Accordingly, we apply a bi-directional GRU network to all the collected candidate representations to encode environments into a unified state representation: 
\begin{flalign}
\scriptsize 
\label{GRU}
& h_t^+ = GRU([\hat{F_0}, \textbf{F}, \textbf{S}]_t, h_{t-1}^+) \\
& h_t^- = GRU([\hat{F_0}, \textbf{F}, \textbf{S}]_t, h_{t+1}^-)   
\end{flalign}
where $h_t^+$ and $h_t^-$ denote the hidden state of the forward and backward GRUs at time $t$, respectively. The bi-directional states $h_t^+$ and $h_t^-$ are concatenated together as the feature vector of unified state representation, i.e., the H = [$h_t^+$, $h_t^-$]. In practice, the collected candidate state representations $[\hat{F_0}, \textbf{F}, \textbf{S}]$, whose total feature dimension is 9218, are transformed into a 1-D feature vector with the dimension of 1024 using GRU.

\textbf{Discussion:} Note that diverse search regions could be more helpful for robust tracking and we have already attempted to maintain and use separate search regions in our implementation. However, by sampling different search regions for different trajectories, we have found that the overall computational complexity, including the costs for feature extraction and information update with respect to the extra search regions, would become extremely large and could make it quite intractable for online tracking. As a result, considering that the TANet module of our method can already provide rich diversified candidate areas for searching, we share the local search regions and global search regions from TANet to avoid excessive complexity and make our method more practical. In addition, regarding the performance of our method, although it is subjected to the baseline method, we would like to mention that our method already improves the baseline methods to a considerable extent, which is usually very difficult when using other methods. This can be illustrated in the experimental comparison. Besides, since our work mainly aims to tackle the multi-trajectory single-object tracking problem based on a novel reinforcement mechanism, we tend to leave the improvement of baseline methods to future works.

\subsubsection{\textbf{Action Selection}}
By taking the unified state representation as input, we define that the action space of each agent is to predict the proposal index that indicates the proposal selected as the tracking result for current frame. We formulate this action space as a continuous action space and normalize the action values into $0\thicksim1$. Then, the true index of selected proposals can be obtained by multiplying the action value with the number of extracted proposals.

A decision making policy network is assigned to each agent to decide which action to take. The network directly predicts the action value. Besides, we also boost communication between decision making policy networks of different agents by devising a novel sequential action selection mechanism. More specifically, in addition to the unified state representation, we also input to each agent the action selected by the previous agent to encourage diversified proposal selection results. For the first agent, we input the action selected by the greedy search mechanism to start the action selection procedure, as shown in Fig. \ref{pipeline}.

\subsubsection{\textbf{Reward Function}}
For the reward function, we take the IoU between selected proposals and the ground truth bounding box as an evaluation metric to encourage the selected proposals to be as close to the ground truth as possible. More specifically, if the agent selects the proposal whose IoU with the ground truth is larger than a pre-defined threshold value, then this agent is rewarded by +1; otherwise, it will be punished by -1. It is also worthy to note that all the agents share the same reward function. Following existing RL algorithms~\cite{mnih2016asynchronous, mnih2015human}, we also utilize accumulated discounted rewards, i.e., the rewards obtained in a more distant future contribute less to the current step. Therefore, the discounted return for each agent at time step $t$ can be formulated as: 
\begin{equation}
\label{returnsFunction}
R_t = \sum_{k=0}^{T-t} \gamma^k r_{k+t}, 
\end{equation}
where $\gamma$ is a discount factor and equal to 0.9 in all our experiments.

\subsection{The Training} 

The objective of our proposed MARL framework is to maximize the expected reward.
Although many reinforcement learning algorithms can be adopted for this task, such as DQN~\cite{mnih2015human} or REINFORCE~\cite{williams1992simple}, in this paper, we adopt the Proximal Policy Optimization Algorithms (PPO)~\cite{schulman2017proximal} to optimize the parameters of bi-directional GRU and policy network due to its stability and efficiency. Compared with regular policy gradient methods, PPO is proposed to force the ratio between the updated and previous policy to stay within an interval which can handle the issue of destructively large policy updates in the learning phase, as noted in~\cite{schulman2017proximal}. Formally, given the action space, the objective of PPO can be expressed as: 
\begin{equation}
\label{ppoObjective}
\small 
J^{CLIP}(\theta) = \mathbb{E} [\min(\rho_t \hat{A}(s, a), clip (\rho_t, 1-\epsilon, 1+\epsilon)  \hat{A}(s, a))], 
\end{equation}
where $\rho_t = \frac{\pi_{\theta} (a_t|s_t)}{\pi_{\theta_{old}} (a_t|s_t)}$ denotes the ratio between two policies, $\pi_{\theta}$ is a stochastic policy and $\pi_{\theta_{old}}$ is the policy before the update. The function $clip (\rho_t, 1-\epsilon, 1+\epsilon)$ constrains the ratio within $[1-\epsilon, 1+\epsilon]$ and $\epsilon$ is a hyper-parameter. $ \hat{A}(s, a)$ is an estimator of the advantage function at time step $t$, which can be obtained by: 
\begin{equation}
\label{advantageFunction} 
 \hat{A}(s, a) = Q(s, a) - V(s), 
\end{equation}
where $V(s)$ is state-value function which measures the expected return of state $s$; $Q(s, a)$ is the action-value function similar to $V(s)$, but it is used to measure the expected return of a pair of state and action $(s, a)$.

In this paper, the PPO is implemented based on an actor-critic architecture that contains two branches, i.e., the actor network (policy function) and critic network (value function). Following existing works~\cite{mnih2016asynchronous, williams1991function, wu2019multiMARL, mnih2015human} which utilize entropy regularization $\mathcal{H}$ to prevent policies from becoming deterministic, we also adopt this technique to encourage \emph{exploration} in the training phase. In other words, the agents are encouraged to select more diverse proposals with larger entropy. Therefore, the learning objective for actor network can be formulated as: 
\begin{equation}
\label{newPPOObjective}
J^{actor}(\theta) =J^{CLIP}(\theta) + c_1 \mathcal{H}(s, \pi_{\theta}(.)), 
\end{equation}
where $c_1$ is a hyper-parameter. 
For the critic network, we compute the mean squared error between the discounted return $R$ and estimated state-value $V(s)$ to update its parameters: 
\begin{equation}
\label{criticObjective}
J^{critic}(\theta) = (R_t - V(s))^2. 
\end{equation}
The overall training process is summarized in \textbf{Algorithm \ref{algorithmALL}}.

\renewcommand{\algorithmicrequire}{\textbf{Input:}}
\renewcommand{\algorithmicensure}{\textbf{Output:}}
\begin{algorithm}
\small 
\caption{Offline Training of the Beam Search Policy.}
\label{algorithmALL}
\begin{algorithmic}[1]
    \REQUIRE  Training videos C and it's ground truth G
    \ENSURE    Beam Search Policy
    
    \STATE    Initialize RT-MDNet with $\theta^{RT}$
    \STATE    Initialize policy and value function parameters with $\theta_0$ and $\phi_{0}$
    \STATE    Initialize the replay buffer D, beam width B  
	
    \STATE    \textbf{for} episode = 1, M \textbf{do}    
    \STATE    ~~\textcolor[RGB]{0,140,0}{$\#~Training~Sample~Collection$}
    \STATE    ~~Randomly select a clip of frames $\{X_k, X_{k+1}, ... , X_{k+T}\}$
    \STATE    ~~Draw samples \{$S^+_1$, $S^-_1$\} to fine-tuning RT-MDNet
    \STATE    ~~\textbf{for} frame t = 2, T+1 \textbf{do}
    \STATE    ~~~~Feed extracted proposals into RT-MDNet to get $state$
    \STATE    ~~~~Select B actions $(a_t^1, a_t^2, ... , a_t^B)$ according to current policy and exploration rate, recurrently
    \STATE    ~~~~Execute B actions, observe reward $r_t$ and next states 
    \STATE    ~~~~Store transitions in D
    \STATE    ~~\textbf{end for}

    \STATE    ~~\textcolor[RGB]{0,140,0}{$\#~Policy~Optimization$}     
    \STATE    ~~Sample mini-batch of N transitions $D_k = \{\tau_i\}$ from D 
    \STATE    ~~Compute the Returns $R_t$ and estimated advantage $\hat{A_t}$ based on current value function $V_{\phi_k}$ 

    \STATE    ~~Update policy by optimizing Eq. (\ref{newPPOObjective}) 
    \STATE    ~~Update value function by optimizing Eq. (\ref{criticObjective}) 
        
    \STATE    \textbf{end for}
            
\end{algorithmic}
\end{algorithm}

\subsection{Beam Search-based Tracking}  
In our proposed beam search-based tracking strategy, multiple tracking results are selected in each frame and will be maintained until processing the whole video sequence. Different from the vanilla beam search mechanism that selects results with top-\emph{k} confidence scores, our strategy selects results from the proposed MARL framework.

Before tracking the target in a test video, we first pre-train the local search tracker following the protocol of MDNet and RT-MDNet and pre-train the global search tracking following TANet. Note that we also follow MDNet and utilize the bounding box regression, short- and long-term update strategy for more accurate tracking. When processing the test video, in each coming frame, we conduct Gaussian sampling on both local and global search regions to obtain candidate state representations and the unified state representation. Then, we feed the state representation and other information as mentioned previously into different agents to select their tracking results. After all the agents have selected their results, we maintain these results and conduct similar operations for subsequent frames until the end of testing video. Once all the video frames are processed, we can obtain multiple tracking trajectories. In this paper, we simply summarize all the scores of the corresponding trajectory and select the best scored one as the tracking result of the current video sequence.

It is worthy to note that it is not intuitive to apply the aforementioned method for the Siamese network based trackers (DiMP~\cite{bhat2019DiMP} is tested in this work), because these trackers directly predict a response map instead of proposals. Hence, we modify the output \emph{action} of each agent as two values, i.e., the coordinates of the selected location in the current response map. We set the actions of previous agent and the response map as the \emph{state} for the subsequent agent and the definition of \emph{reward} remains the same.

\section{Experiment}

\subsection{Datasets and Evaluation Metric} 
Our method is effective for both short-term and long-term videos. In this paper, we use both the short-term and long-term videos for training (including \textbf{TLP}~\cite{moudgil2018long} and \textbf{DTB}~\cite{li2017visualuav}), which  totally contain 120 video sequences. Note that, some of the long-term videos consist of more than 20K frames.
The following tracking datasets are used for inference and compare with other state-of-the-art trackers: \textbf{OTB2015}~\cite{wu2015object}, \textbf{TC128}~\cite{Liang2015Encoding}, \textbf{UAV123}~\cite{mueller2016benchmarkuav20l}, \textbf{LaSOT}~\cite{fan2019lasot}, \textbf{GOT-10K}~\cite{huang2018got}, \textbf{VOT2018-LT}~\cite{kristanvot2018lt}, and \textbf{VOT2019-LT}~\cite{kristanvot2019lt}.

\textbf{PR} (Precision Rate) and \textbf{SR} (Success Rate) are two metrics widely used in the tracking community. The first evaluation metric illustrates the percentage of frames where the center location error between the object location and ground truth is smaller than a pre-defined threshold (20 pixel threshold are usually adopted). The second one demonstrates the percentage of frames the Intersection over Union (IoU) of the predicted and the ground truth bounding boxes is higher than a given ratio. In addition, the \textbf{AO} (Average Overlap) is also used for the evaluation of GOT-10K dataset which denotes the average of overlaps between all ground truth and estimated bounding boxes. As noted in~\cite{huang2018got}, the AO is recently proved to be equivalent to the area under curve (AUC) metric employed in OTB-100~\cite{wu2015object} and LaSOT~\cite{fan2019lasot}. For the VOT2018-LT and VOT2019-LT datasets, we adopt their default metrics as the evaluation criteria, i.e., \textbf{Precision}, \textbf{Recall}, and \textbf{F1-score}. Specifically, the definition of these metrics are: 
\begin{equation}
\label{Precision}
\small  
Precision = \frac{TP}{TP + FP}, ~~~~Recall = \frac{TP}{TP + FN}, 
\end{equation}
\begin{equation}
\label{Precision}
\small  
F1-score =  \frac{2 \times Precision \times Recall}{Precision + Recall}, 
\end{equation}
where TP, FP and FN are used to denote the True Positive, False Positive and False Negative, respectively.

\subsection{Implementation Details}

For the training of TANet, the batch size is 20, the initial learning rate is 0.0004, and the maximum epoch is 15. The ground truth mask used for the training is obtained by filling the target and background regions with white and black pixels. 
For the beam search policy network, the learning rate for the actor network and critic network are 0.001 and 0.005, respectively, entropy weight $c_1$ is set as 0.005, $\gamma$ is 0.9, $\tau$ is 0.8. The input dimension of bi-directional GRU is 9218 and the encoded hidden state is 1024. It is also worthy to note that the pre-trained RT-MDNet~\cite{Jung_2018_ECCV}\footnote{\url{https://github.com/BossBobxuan/RT-MDNet}} is utilized in the training phase of our beam search policy network due to its high efficiency. But the learned beam search policy also works well when integrating with MDNet~\cite{Nam2015Learning}\footnote{\url{https://github.com/HyeonseobNam/py-MDNet}}  according to our experimental results. To validate the generalization of our proposed beam search strategy, we also integrate it with strong tracker DiMP~\cite{bhat2019DiMP} to check the final results. 
To finish training, the beam search network of our method requires about 7 days to find a better checkpoint for testing using a server with eight RTX2080TI GPUs.
The code will be released at \url{https://github.com/wangxiao5791509/BeamTracking}.

\subsection{Comparison on Public Benchmarks}
In this subsection, we report our tracking results and compare with other state-of-the-art trackers on seven popular tracking benchmark datasets.  

\begin{table*}[!htp]
\center
\scriptsize 
\caption{Tracking results on OTB-2015, TC-128, UAV123, GOT-10K, VOT2018-LT, and VOT2019-LT dataset.} \label{OTBUAVTCresults}
\resizebox{\textwidth}{25mm}{
\begin{tabular}{c|ccccccc|cc}
\hline \toprule [0.7 pt]
\rowcolor{mygray}     
\textbf{Algorithm}    &PGNet~\cite{liao2020pgnet}   &Ocean ~\cite{zhangocean}   &PTAV~\cite{fan2017ptav}      &SiamRCNN~\cite{2020siamRCNN}          &CREST~\cite{song-iccv17-CREST}			  &TANet~\cite{wang2020ganTANetTrack}        		&GCT~\cite{gao2019GCT}				&MDNet~\cite{Nam2015Learning}   		&Ours     \\ 
\textbf{OTB-2015 (PR|SR)}  &0.892|0.691   &{0.920}|0.684  &0.848|0.634   &0.891|{0.701}    &0.838|0.623     &0.791|0.646       	&0.854|0.648	&0.868|0.645     	&0.886|0.653       	\\   
\hline \toprule [0.7 pt]
\rowcolor{mygray}
\textbf{Algorithm} 	 &GradNet~\cite{li2019gradnet}   &PTAV~\cite{fan2017ptav}  &ACT~\cite{chen2018realACT}   &MEEM ~\cite{zhang2014meem}   &SINT~\cite{Tao2016Siamese}    &ADNet ~\cite{Yun2017Action}  &RT-MDNet ~\cite{Jung_2018_ECCV}   &MDNet~\cite{Nam2015Learning}   &Ours    \\
\textbf{TC-128  (PR|SR)}  &0.764|0.556   &0.741|0.544   &0.738|0.532   &0.675|0.483    &0.711|0.521    &0.761|0.558    &0.767|0.559    &0.797|0.569    &0.806|0.583    \\
\hline \toprule [0.7 pt]
\rowcolor{mygray}
\textbf{Algorithm}	&DTNet~\cite{zhang2020DTNet}    &SAMF~\cite{li2014samf}  &SRDCF~\cite{danelljan2015learning}  &ECO~\cite{Danelljan2016ECO}    &SiamRPN~\cite{li2018siamRPN}   &DaSiamRPN~\cite{zhu2018distractor}      &THOR~\cite{sauer2019thor}       &MDNet ~\cite{Nam2015Learning} &Ours       \\
\textbf{UAV123  (PR|SR)}		& 0.726|0.539    &0.592|0.396      & 0.676|0.464     &0.741|0.525   &0.748|0.527     &0.796|0.586      &0.758|0.697        &0.747|0.528      &0.773|0.668    \\
\hline \toprule [0.7 pt]
\rowcolor{mygray}
\textbf{Algorithm}  &ATOM~\cite{danelljan2019atom}	& GOTURN~\cite{held2016GOTURN}	& SiamFC~\cite{Bertinetto2016SiameseFC}  & SiamFC++~\cite{xu2020siamfc++} &Ocean~\cite{zhangocean}  &KYS~\cite{bhat2020KYS}  	&D3S~\cite{lukezic2020d3s} & DiMP~\cite{bhat2019DiMP}  	 & Ours	\\ 	
\textbf{GOT-10K ($AO|SR_{0.5}$)} 		&0.556|0.634	&0.347|0.375	&0.348|0.353		&0.595|0.695   &0.611|0.721   &0.636|0.751   &0.597|0.676  &0.673|0.785		&0.685|0.800 		\\ 
\hline \toprule [0.7 pt]
\rowcolor{mygray}
\textbf{VOT2018-LT} 	&PTAV+~\cite{fan2017ptav}  &SYT~\cite{kristanvot2018lt}  &SINT-LT~\cite{Tao2016Siamese}  &MMLT~\cite{lee2018MMLT}  &DaSiam-LT~\cite{zhu2018distractor}     &MBMD~\cite{zhang2018learningTermTracking}   &SiamRPN++~\cite{li2018siamrpn++}   &DiMP~\cite{bhat2019DiMP} &Ours \\
Precision 		&0.595     &0.520      &0.566      &0.574    &0.627      					&0.634       			&0.646       &0.660       &0.683        \\
Recall 			&0.404     &0.499      &0.510      &0.521    &{0.588}      				&{0.588}       &0.419       &0.583       &0.613        \\
F1-score 		&0.481     &0.509      &0.536      &0.546    &0.607     				 	&{0.610}       &0.508        &0.619       &0.646        \\
\hline \toprule [0.7 pt]
\rowcolor{mygray}
\textbf{VOT2019-LT} 	 &ASINT~\cite{kristanvot2019lt}  &CooSiam ~\cite{kristanvot2019lt} &Siamfcos-LT ~\cite{kristanvot2019lt} &SiamRPNs-LT~\cite{li2018siamRPN}     &mbdet~\cite{kristanvot2019lt}   &SiamDW-LT~\cite{zhipeng2019deeper}  &SiamRPN++~\cite{li2018siamrpn++} &DiMP~\cite{bhat2019DiMP}  &Ours \\
Precision 		 &0.520      &0.566      &0.574    &0.627      &0.634       &{0.649}      &0.627    &0.655       &0.679        \\
Recall 			 &0.499      &0.510      &0.521    &0.588      &0.588       &{0.609}      &0.399   &0.572       &0.601        \\
F1-score 		&0.509      &0.536      &0.546    &0.607      &0.610        &{0.629}      &0.488   &0.611       &0.638        \\
\hline \toprule [0.7 pt]
\end{tabular}  }
\end{table*}

\textbf{Results on OTB-2015~\cite{wu2015object}:} 
OTB-2015 is the first large-scale standard benchmarks for visual tracking proposed by Wu et al. It contains 100 video sequences which reflect 9 challenging attributes, such as \emph{illumination variation, scale variation, occlusion, deformation}. We compare our tracker with PGNet~\cite{liao2020pgnet}, Ocean~\cite{zhangocean}, PTAV~\cite{fan2017ptav}, SiamRCNN~\cite{2020siamRCNN}, CREST~\cite{song-iccv17-CREST}, TANet~\cite{wang2020ganTANetTrack}, GCT~\cite{gao2019GCT}, and MDNet~\cite{Nam2015Learning} on this benchmark. From the Table \ref{OTBUAVTCresults}, we can find that the baseline method MDNet achieves $0.868|0.645$ on the OTB-2015 tracking benchmark. 
Our approach obtains $0.886|0.653$ on this benchmark, which outperforms the baseline method by $+1.8\%, +0.8\%$ on both metrics respectively. In addition, our method is also better than other compared tracking algorithms, such as PTAV ($+3.8\%, +1.9\%$), CREST ($+4.8\%, +3.0\%$), TANet ($+9.5\%, +0.7\%$), and GCT ($+3.2\%, +0.5\%$).

\textbf{Results on TC128~\cite{Liang2015Encoding}:}  
TC128 is specifically designed for the evaluation of color related trackers which contains 129 testing video sequences. 
As shown in Table \ref{OTBUAVTCresults}, the baseline method MDNet achieves $0.797|0.569$ on the PR/SR. In contrast, our proposed beam search policy achieves better results on this benchmark ($0.806|0.583$) than baseline method and other compared trackers, including GradNet~\cite{li2019gradnet}, PTAV~\cite{fan2017ptav}, ACT~\cite{chen2018realACT}, MEEM~\cite{zhang2014meem}, SINT~\cite{Tao2016Siamese}, ADNet~\cite{Yun2017Action}, and RT-MDNet~\cite{Jung_2018_ECCV}. 

\textbf{Results on UAV123~\cite{mueller2016benchmarkuav20l}:}   
UAV123 is a dataset specifically designed for UAV tracking which contains 123 video sequences. From Table \ref{OTBUAVTCresults}, the MDNet achieves $0.747|0.528$ on the PR|SR, while our tracker improve them to  $0.773|0.668$. Our tracker also outperforms the DTNet~\cite{zhang2020DTNet}, SiamRPN~\cite{li2018siamRPN}, and DaSiamRPN~\cite{zhu2018distractor}.

\textbf{Results on GOT-10K~\cite{huang2018got}:} 
GOT-10K is constructed based on the backbone of WordNet structure~\cite{miller1995wordnet}. It populates the majority of over 560 classes of moving objects and 87 motion patterns. It contains 10,000 videos totally, with more than 1.5 million manually labeled bounding boxes. The authors select 280 videos as the test subset and the rest of videos are used for training. As we can see from Table \ref{OTBUAVTCresults}, our tracker achieves better results than the baseline method DiMP~\cite{bhat2019DiMP}. Specifically speaking, the DiMP achieves $0.673|0.785$ on the $AO|SR_{0.50}$, respectively. When integrating the multi-agent beam search strategy, our tracker obtained $0.685|0.800$ on the two metrics, respectively. Compared with other trackers including ATOM~\cite{danelljan2019atom}, GOTURN~\cite{held2016GOTURN}, SiamFC~\cite{Bertinetto2016SiameseFC}, SiamFC++~\cite{xu2020siamfc++}, Ocean~\cite{zhangocean}, KYS~\cite{bhat2020KYS}, and D3S~\cite{lukezic2020d3s}, our overall tracking performance is also better than theirs. 

\begin{figure*} 
\center
\includegraphics[width=7in]{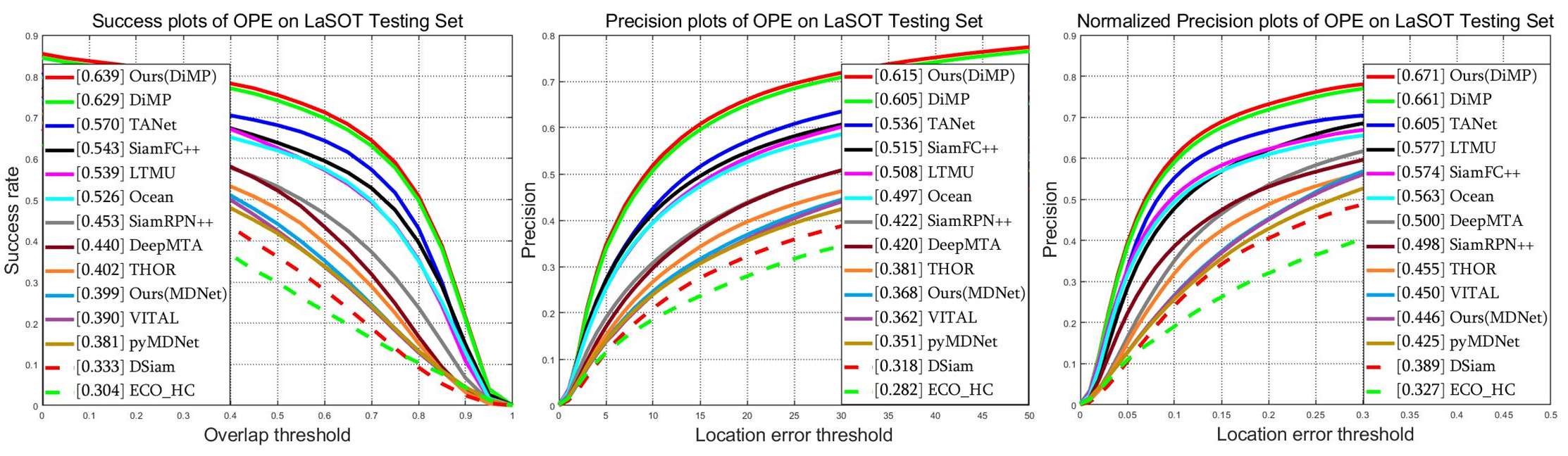}
\caption{Tracking results on LaSOT dataset.}
\label{LaSOTresults}
\end{figure*}

\textbf{Results on LaSOT~\cite{fan2019lasot}:} 
LaSOT is the currently largest long-term tracking dataset which contains 1400 video sequences with more than 3.5M frames in total. The average video length is more than 2,500 frames and each video contains challenging factors deriving from the wild, e.g., out-of-view, scale variation. It provides both natural language and bounding box annotations which can be used for the explorations of integrating visual and natural language features for robust tracking. For the evaluation of LaSOT dataset, we test our tracker based on the \emph{Protocol  II} which contains 280 videos. 
As shown in Fig. \ref{LaSOTresults}, the baseline method MDNet achieves $0.351|0.381$ on the PR and SR respectively on the LaSOT benchmark dataset. Our proposed tracker achieves better results than the baseline, i.e., $0.368|0.399$ on the two evaluation metrics respectively. When integrated with strong trackers like DiMP, our proposed MARL based beam search strategy can also improve over this baseline. Specifically, the DiMP achieves $0.605|0.661|0.629$ on Precision Plot, Normalized Precision Plot and Success Plot, while we attain $0.615|0.671|0.639$ respectively. The experiments on this benchmark demonstrate the effectiveness and generalization of our proposed modules for tracking. Compared with other state-of-the-art trackers, such as SiamFC++~\cite{xu2020siamfc++}, LTMU~\cite{dai2020LTMU}, Ocean~\cite{zhangocean}, and TANet~\cite{wang2020ganTANetTrack}, our results are also better than them. 

\textbf{Results on VOT2018-LT~\cite{kristanvot2018lt}: }  
VOT2018-LT is a long-term dataset which contains 35 videos with a total length of 146,817 frames. It is calculated that the target object will disappear for 12 times and each lasting on average 40 frames for each video.
As shown in Table \ref{OTBUAVTCresults}, the baseline DiMP achieves 0.660, 0.583, 0.619 on the Precision, Recall and F1-score, while our method achieves 0.683, 0.613, 0.646, respectively. In addition, our tracker also  outperforms other compared trackers by a large margin. 

\textbf{Results on VOT2019-LT~\cite{kristanvot2019lt}: }
VOT2019-LT totally contains 50 videos with 215,294 frames and each video contains on average 10 long-term target disappearances. Based on VOT2018-LT, this dataset introduces external 15 videos which are more challenging. 
We also evaluate our tracker on a more challenging VOT2019-LT dataset in Table \ref{OTBUAVTCresults}. The baseline tracker DiMP attains 0.655, 0.572, 0.611, while we achieve 0.679, 0.601, 0.638 on Precision, Recall and F1-score, respectively. Compared with other long-term trackers, our results are also better than them which demonstrate the advantages of our proposed MARL beam search strategy.

All in all, the aforementioned experiments fully validated that our MARL based beam search strategy is effective for visual tracking.

\subsection{Ablation Study}

\subsubsection{Component Analysis} 
In this section, we conduct ablation studies on the GOT-10K and VOT-2016 dataset. The MDNet and RT-MDNet are adopted as the baseline tracker respectively. Various search strategies are discussed as described below: 
\begin{itemize}
\item \textbf{{Vanilla Greedy Search (VGS):}}	This inference approach is widely used in existing visual trackers. We take this search strategy as our baseline method. 

\item  \textbf{{Global Search (GS):}}	We take the global proposal into baseline tracker to achieve joint local and global search. 

\item  \textbf{{Single-Agent Greedy Search (SAGS):}} Following greedy search used in popular tracking algorithms, we train a greedy search policy with single agent based reinforcement learning. Then, we integrate it into the baseline tracker to discuss the difference between greedy select policy and our learning-based select policy. 

\item  \textbf{{Na\"ive Beam Search (NBS):}} We adopt beam search algorithm for tracking, which target at selecting and maintaining the top-$k$ proposals for each video frame according to the classification score. 

\item  \textbf{{Multi-Agent Beam Search (MABS):}} Following the idea of beam search algorithm, we utilize our multi-agent reinforcement learning based beam search policy for visual tracking. 
\end{itemize}

\begin{table}[!htp]
\center
\scriptsize 
\caption{Component analysis on the GOT-10K and VOT-2016. AO|SR and AUC score are reported on the two datasets, respectively.} \label{GOT10kablationStudy}
\begin{tabular}{c|ccccc|c|cccccc}
\hline \toprule [0.7 pt]
Index   								&VGS		&GS	  &SAGS 		&NBS 		&MABS 	&GOT-10K   &VOT-2016 \\ 	
\hline 
\ding{172} 				&\cmark			&						&						&			&						&$0.299|0.303$	&$0.644$				\\
\ding{173}				&\cmark			&\cmark			&						&			&						&$0.381|0.402$	&$0.692$				\\
\ding{174}				&						&\cmark			&\cmark			&			&						&$0.383|0.407$	&$0.723$				\\
\ding{175}				&						&\cmark			&						&\cmark			&			&$0.399|0.439$	&$0.710$				\\
\ding{176}				&						&\cmark			&						&			&\cmark			&$0.412|0.449$	&$0.730$				\\
\hline \toprule [0.7 pt]
\end{tabular}
\end{table}

As shown in Table \ref{GOT10kablationStudy}, we can find that the baseline method which adopts Vanilla Greedy Search for visual tracking achieves $0.299|0.303$ on AO|SR, respectively. When combining the TANet model with baseline method for joint local and global search, the tracking performance can be significantly improved to $0.381|0.402$. This experiment fully validated the effectiveness of our introduced TANet. 
When integrating with our single agent based search strategy, i.e., MDNet+GS+SAGS, we can further improve the result to $0.383|0.407$. This result fully validated the effectiveness of our proposed non-greedy search algorithm (Comparing with regular classifier based trackers which always select the proposal with maximum score, our tracker select the proposal our agent recommended). 
On the basis of this implementation, we propose a novel beam search strategy for visual tracking based on MARL. As shown in Table \ref{GOT10kablationStudy}, we can find that our beam search achieves the best tracking performance on this benchmark compared with other search strategies, including greedy search, single agent search. 
It is also worthy to note that our MARL based beam search ($0.412|0.449$) also achieves better results than na\"ive beam search policy ($0.399|0.439$). It is because the na\"ive beam search policy only select the top-$k$ proposal for each frame, however, these proposals sometimes are nearly the same. Our beam search policy select the proposals recommended by various agents which will get rid of such $\emph{greedy}$ property,   therefore, we can obtain better results.

In addition to aforementioned analysis, we also combine these search strategies with RT-MDNet and test them on the VOT-2016 dataset. Due to the fact that our proposed approach have multiple tracking results for each frame, therefore, the default evaluation metric of VOT-2016 is not suitable for the evaluation. In this experiment, we adopt the AUC score as the evaluation metric. From Table \ref{GOT10kablationStudy}, we can draw similar conclusions with the experimental results based on MDNet. Specifically, compared with na\"ive beam search, our MARL based beam search scheme obtain $+2 \%$ points improvement. These experimental results fully validate the effectiveness and advantages of our proposed MARL based beam search scheme for visual tracking task.

\subsubsection{Analysis on Number of Trajectories} 
\begin{figure}[!htp]
\center
\includegraphics[width=3in]{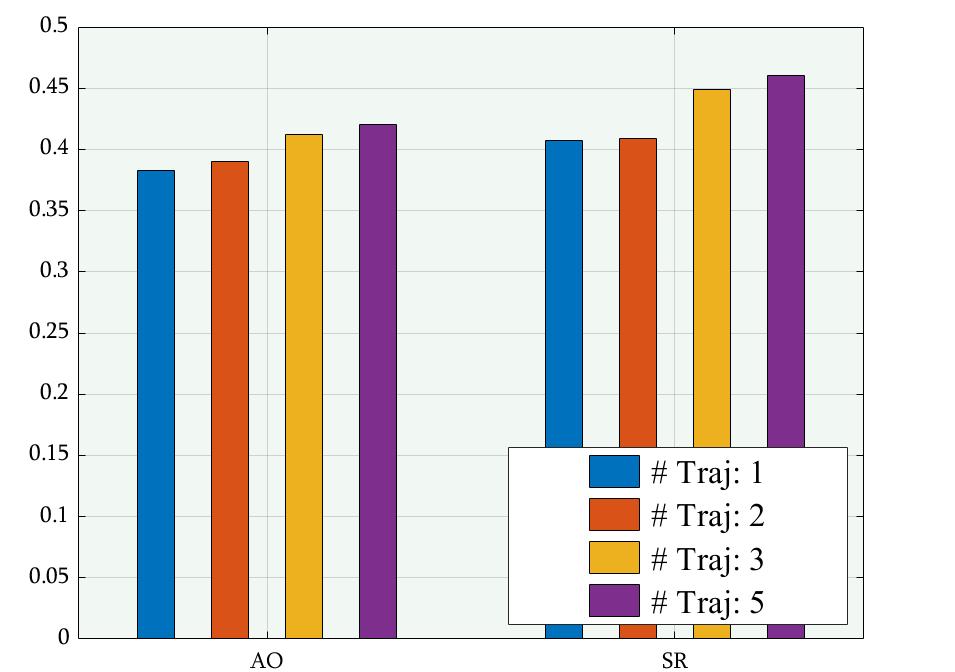}
\caption{Results with various number of trajectories on GOT-10k dataset. }
\label{trajNUManalysis}
\end{figure}

As shown in Fig. \ref{trajNUManalysis}, it's easy to find that our model can obtain better results when increasing the beam width, i.e., the number of tracking trajectories. Specifically, we can obtain $0.383|0.407$ on PR|SR if the beam width is 1; when the beam width is increased to 2, 3, 5, we can attain $0.390|0.409$, $0.412|0.449$, $0.420|0.460$, respectively. 

\subsubsection{Attribute Analysis}
\begin{figure*}[!htp]
\center
\includegraphics[width=7in]{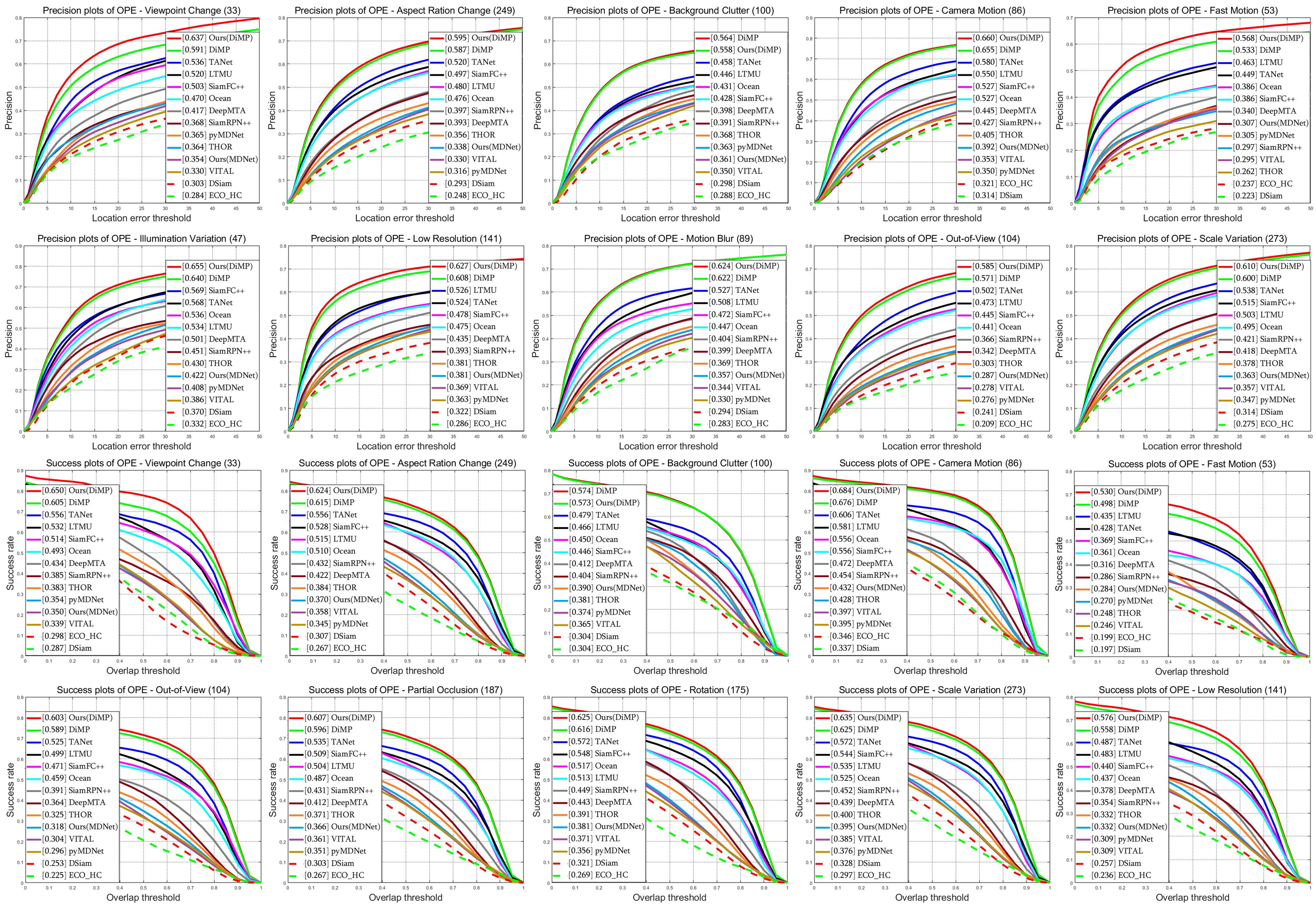}
\caption{Attribute analysis of PR and SR on the LaSOT benchmark. Best viewed by zooming in. }
\label{lasotResultsAAPR}
\end{figure*}

In this paper, we analyse the attribute results of our tracker based on LaSOT tracking benchmark. As shown in Fig. \ref{lasotResultsAAPR}, it is intuitive to find that our tracker is robust to many challenging factors, such as \emph{out-of-view, rotation, camera motion, low resolution} and \emph{fast motion}, compared with other tracking algorithms. This experiment fully validated the effectiveness and robustness of our tracker when tracking in complex scenarios. 

\subsection{Efficiency Analysis}
When integrating our proposed search scheme into MDNet and RT-MDNet, we can achieve 0.67 and 4.27 FPS, respectively. Although our tracker achieves lower running efficiency based on the two trackers, however, our tracker can improve corresponding tracking algorithms significantly on multiple benchmark datasets. When combining it with DiMP tracker~\cite{bhat2019DiMP}, our tracker can run at 22.65 FPS, meanwhile, the overall results can also be improved.

\subsection{Discussion} 
We compare our proposed algorithms with other state-of-the-art trackers on multiple benchmark datasets in previous sections. In this section, we focus on the comparison with RL-based, ensemble learning-based trackers, and hyperparameter optimization. 

\textbf{Compare with RL based trackers:~~~} 
In the experiments, we compare our tracker with other RL based tracking algorithm, including ADNet~\cite{Yun2017Action}, ACT~\cite{chen2018realACT}, DTNet~\cite{zhang2020DTNet}. Specifically, ADNet achieves $0.903|0.659$, $0.880|0.646$ on OTB-2013 and OTB-2015 dataset, respectively and our results are all better than theirs, i.e., $0.940|0.686$, $0.886|0.653$. On the TC128 dataset, we also outperform the ACT which is developed based on MDNet and RL (ACT: $0.738|0.532$, Ours: $0.806|0.583$). DTNet is developed by Song et al. for adaptive switch between detection and tracking based on a hierarchical RL algorithm published in Neurips-2020. We also outperform this tracker (Integrated version: ACT+FCT+SiamFC) on UAV123 dataset, i.e., $0.773|0.668$ vs $0.726|0.539$, on PR|SR respectively. From all the comparisons with RL trackers, we can find that our proposed MARL-based beam search strategy indeed achieves better results and this fully demonstrates the leading performance in the family of single object tracking algorithms based on RL.

\textbf{Compare with ensemble learning based trackers:~~~} 
The ensemble learning based trackers are also related to our proposed algorithms. The ensemble trackers reviewed in related work are used for the comparison, for example, MTA~\cite{lee2015multihypothesis}, DeepMTA~\cite{wang2021deepmta}, MEEM~\cite{zhang2014meem}, CF2~\cite{ma2015hierarchical}, HDT~\cite{qi2016HDT}, MCCT~\cite{wang2018multicueTrack}, Xie et al~\cite{xie2019multiFusion}. Specifically, the MTA achieves $0.838|0.595$  on the OTB-2013 dataset, while we get $0.940|0.686$ which are significantly better than theirs. Our results are also comparable with MCCT~\cite{wang2018multicueTrack} ($0.928|0.714$) on this benchmark. The DeepMTA~\cite{wang2021deepmta} further extends MTA with deep neural networks and achieves $0.799|0.650$ on the OTB-2015 dataset, in contrast, we can attain $0.886|0.653$ on this benchmark dataset. Our results are also better than CF2~\cite{ma2015hierarchical} ($0.837|0.562$), HDT~\cite{qi2016HDT} ($0.848|0.564$). Xie et al.~\cite{xie2019multiFusion} ($0.927|0.873$) outperform ours on this benchmark, however, their tracker needs multiple trackers for final fusion which may need more memory. In addition, their results on large-scale benchmarks are still unknown, in contrast, we can achieve good performance on both short and long-term tracking datasets. On the TC-128 dataset, the MEEM~\cite{zhang2014meem} achieve $0.675|0.483$, while we can attain $0.806|0.583$. Therefore, according to the aforementioned analysis and comparison, we can find that our results are better than most of the ensemble trackers. These experimental results fully demonstrate the advantages over existing ensemble trackers.

\begin{figure*}[!htp]
\center
\includegraphics[width=7in]{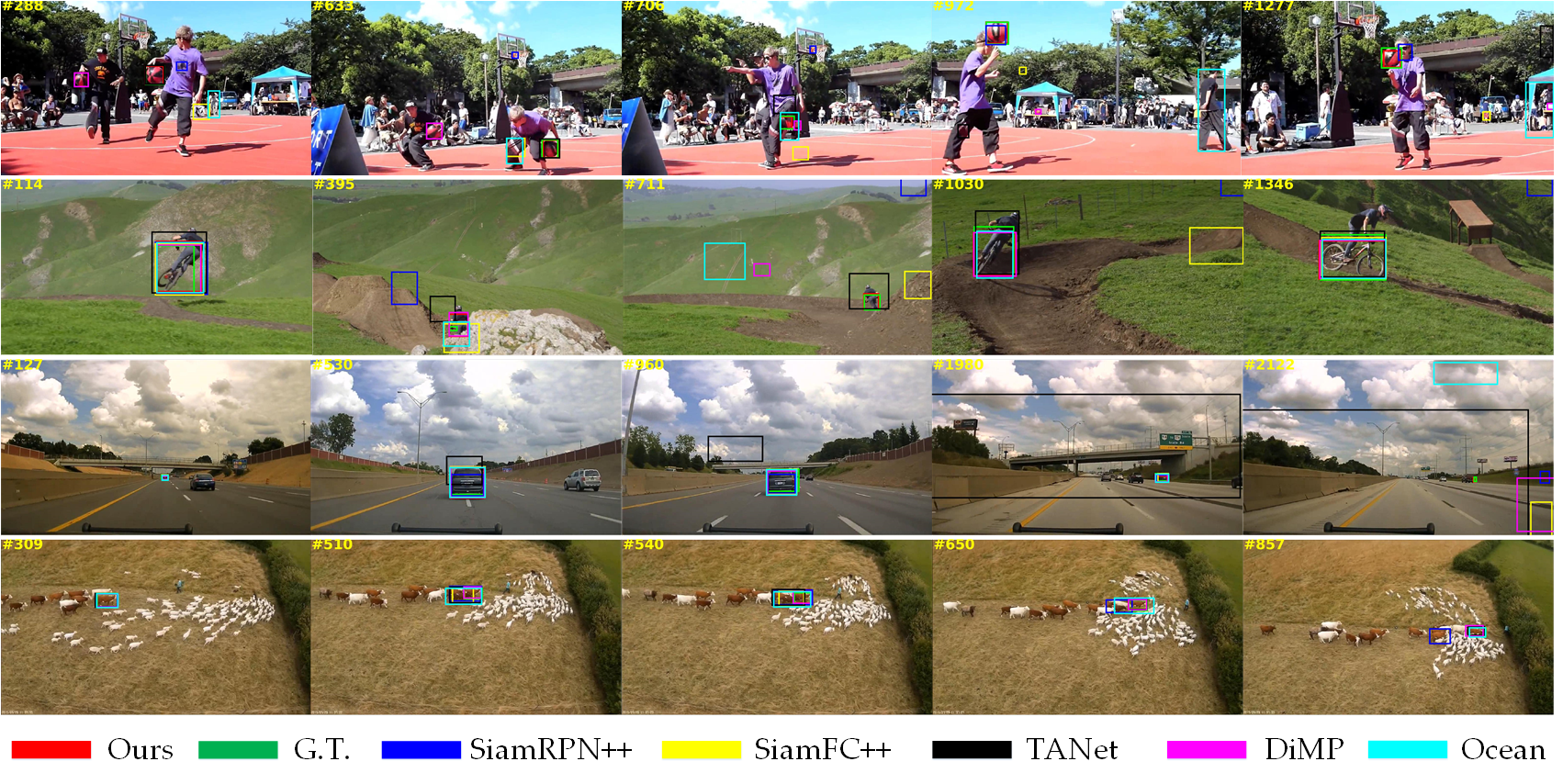}
\caption{Visualization of tracking results of ours and the compared trackers. The corresponding demo videos can be found in our project page.} 
\label{resultVis_lasot}
\end{figure*}

\textbf{Hyperparameter optimization:~~~} 
As is known to all, the hyperparameters are very important for the machine learning models, of course, it is also very important for the visual object tracking task. Usually, these hyperparameters are manually tuned using the control variates approach, which is time-consumption and very challenging to find a set of suitable parameters. Therefore, some automatic hyperparameter tuning methods drawing more and more attention in recent years. The researchers also find that the popular Siamese trackers are sensitive to parameters, and part of the reason they get strong results is because of the hyperparametric tuning tool~\cite{fu2021stmtrack, zhangocean, zhang2021learn}. Different from the above method of brute force search, Dong et al.~\cite{dong2019dynamical} propose the reinforcement learning-based dynamic parameter tuning which is more intelligent. In this paper, we aim to propose the first MARL-based beam search framework for visual object tracking and get top-notch results is not the most important goal. We empirically set the hyperparameters for our tracker, in another word, no multiple parameter adjustments were made to obtain better results. However, reinforcement learning-based dynamic parameter tuning is interesting and useful and can be used to further improve this work. We consider leaving it as our future work.

\subsection{Visualization}
In this section, we demonstrate some illustrations of tracking results on the LaSOT benchmark dataset in Fig. \ref{resultVis_lasot}. It is easy to find that our tracker can obtain better results in challenging environments than the compared method, including SiamRPN++~\cite{li2018siamrpn++}, SiamFC++~\cite{xu2020siamfc++}, TANet~\cite{wang2020ganTANetTrack}, DiMP~\cite{bhat2019DiMP} and Ocean~\cite{zhangocean}. It is also worthy to note that the TANet~\cite{wang2020ganTANetTrack} also adopts a local-global search scheme for tracking, however, it sometimes loses the target object. For example, the black rectangle (i.e., the TANet) fails to locate the \emph{bike} in frame 395 (second row). It also predicts the wrong scale of the target object like the \emph{car} in the third row. These qualitative analyses fully demonstrate the good performance of our tracker which is developed based on MARL based beam search scheme. More tracking results can be found in our demo video on the project page.

\subsection{Failed Cases} 
\begin{figure}[!htp] 
\center
\includegraphics[width=3.5in]{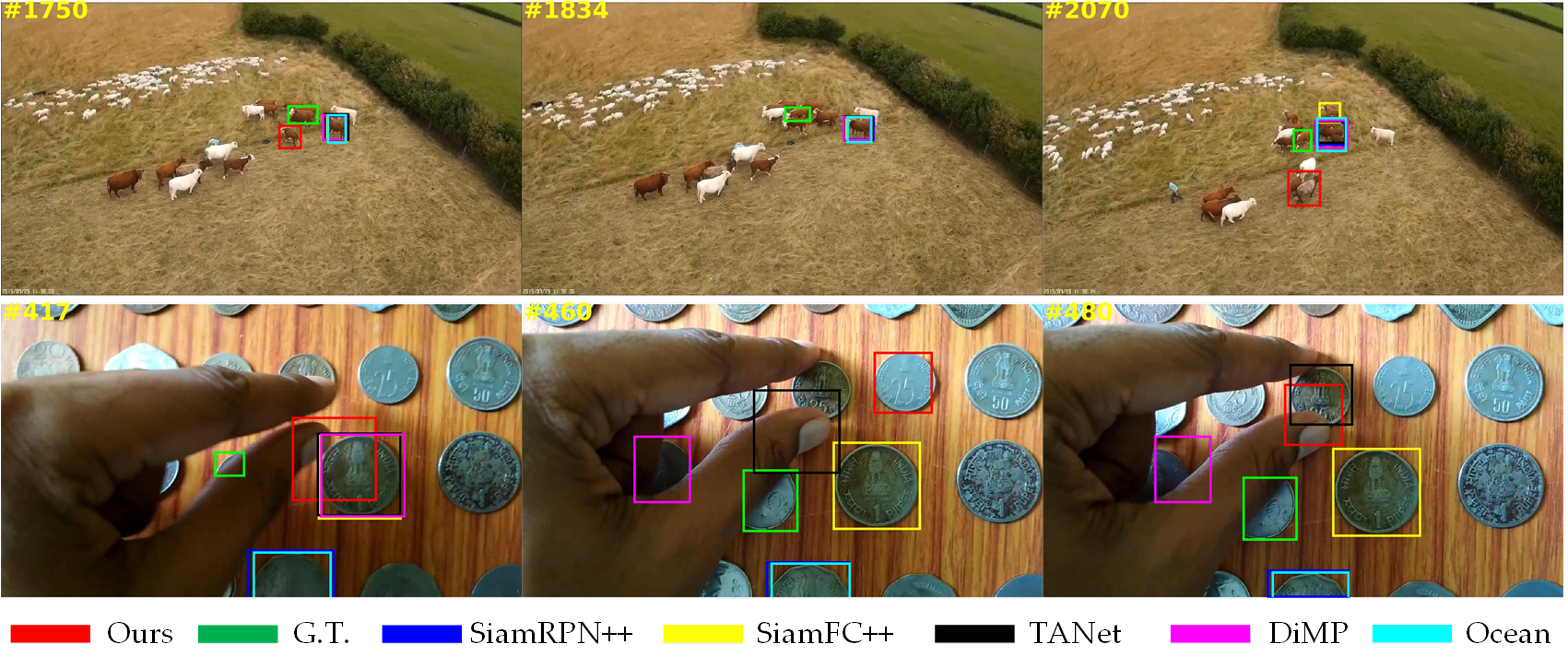}
\caption{Failed cases of our tracking algorithm in complex scenarios.} 
\label{failedCases}
\end{figure}

Although our tracker attains better results on many video sequences, however, it still suffers from the challenging factors, like \emph{small object with dense distractors}, as shown in Fig. \ref{failedCases}. Due to the dense background objects, our tracker fails to locate the right object, for example, the \emph{cow} and \emph{coin} in the first and second row, respectively. In our future works, we will consider adopting graph matching algorithms to model the temporal relations of the target object and background distractors between consecutive frames to help address this issue.  

\section{Conclusion and Future Works}
In this paper, we propose a novel beam search policy for visual tracking based on multi-agent deep reinforcement learning. We attempt to select multiple candidate locations as tracking results of the current frame instead of only one location predicted by a greedy search. We formulate the multiple candidate location selections as a multi-agent decision-making problem and optimize them with PPO algorithm. We also introduce the global search module TANet to handle the re-detection problem in the long-term tracking task. Extensive experiments on multiple tracking benchmark datasets fully demonstrate the effectiveness and generalization of our beam search strategy for tracking.  
In the future works, we will consider modeling the relations between the target object and the background objects to handle the issue of similar target objects for more accurate tracking. We will also exploit the adaptive hyperparameter tuning for high-performance tracking.

\section*{Acknowledgement}
This work is supported by National Natural Science Foundation of China (No. 62102205, 62076003), the University Synergy Innovation Program of Anhui Province (GXXT-2021-038), Major Project for New Generation of AI under Grant (No. 2018AAA0100400), Anhui Provincial Key Research and Development Program (2022i01020014). Dr Zhe Chen was supported by IH180100002.

\bibliographystyle{IEEEtran}
\bibliography{reference}

\vspace{-0.5cm}
\begin{IEEEbiography} [{\includegraphics[width=1in,height=1.25in,clip,keepaspectratio]{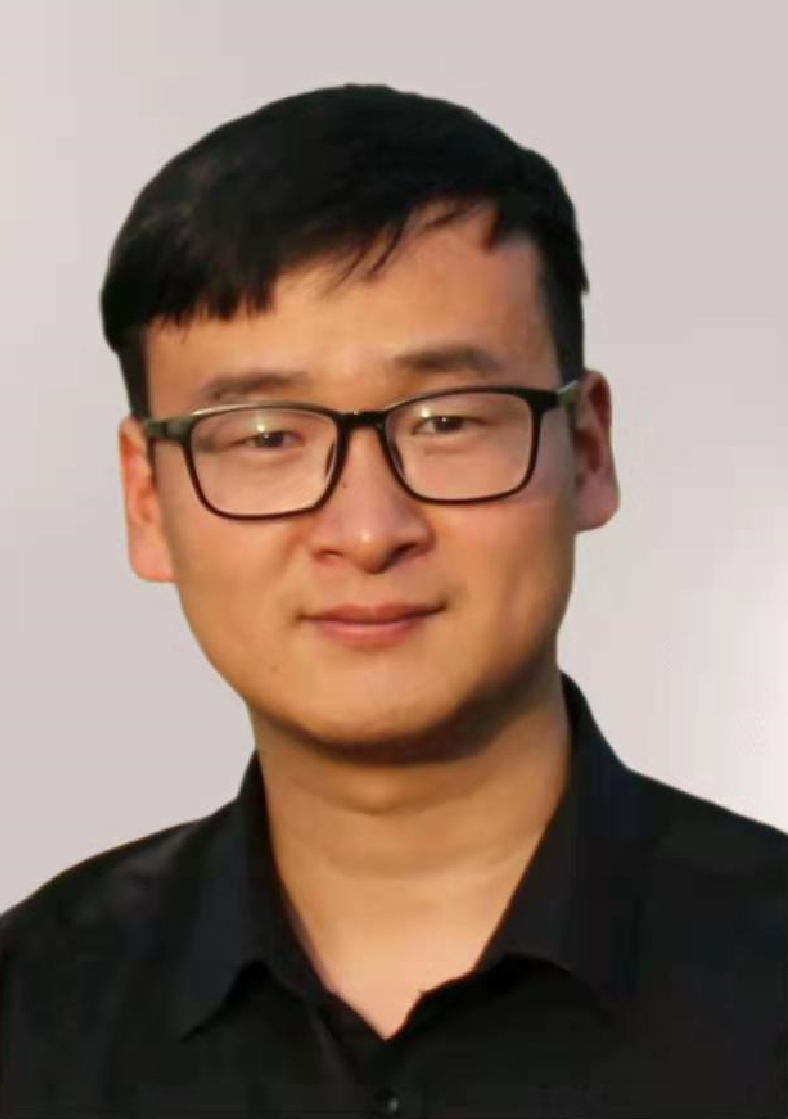}}]
{Xiao Wang} (Member, IEEE) received the B.S. degree in West Anhui University, Luan, China, in 2013. He received the Ph.D. degree in computer science in Anhui University, Hefei, China, in 2019. From 2015 and 2016, he was a visiting student with the School of Data and Computer Science, Sun Yat-sen University, Guangzhou, China. He also has a visiting at UBTECH Sydney Artificial Intelligence Centre, the Faculty of Engineering, the University of Sydney, in 2019. He finished the postdoc research in Peng Cheng Laboratory, Shenzhen, China, from April, 2020 to April, 2022. He is now an Associate Professor at School of Computer Science and Technology, Anhui University, Hefei, China. His current research interests mainly about Computer Vision, Event-based Vision, Machine Learning, and Pattern Recognition. He serves as a reviewer for a number of journals and conferences such as IEEE TCSVT, TIP, IJCV, CVIU, PR, CVPR, ICCV, AAAI, ECCV, ACCV, ACM-MM, and WACV. 
\end{IEEEbiography}

\vspace{-0.5cm}
\begin{IEEEbiography} [{\includegraphics[width=1in,height=1.25in,clip,keepaspectratio]{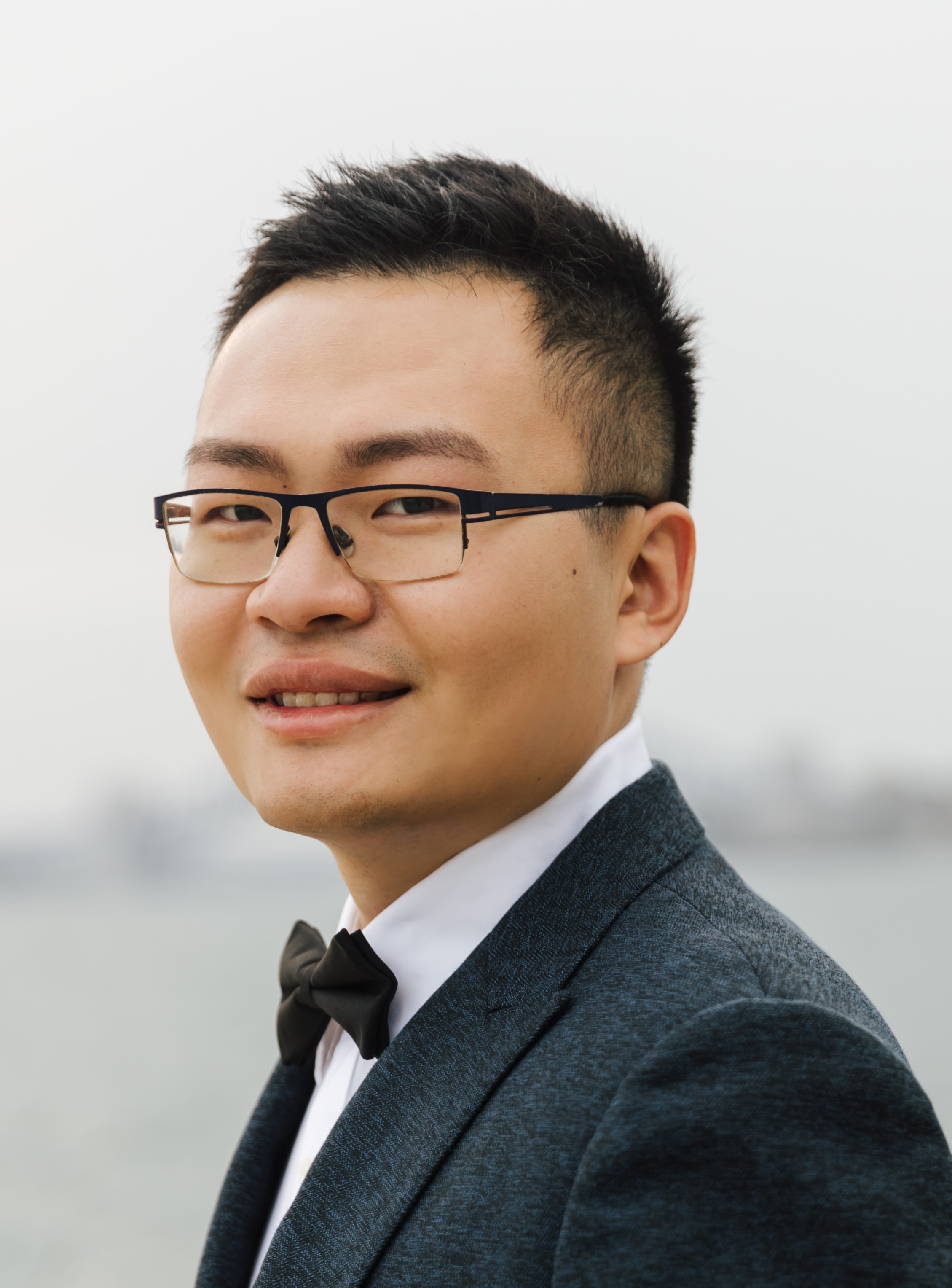}}]
{Zhe Chen} (Member, IEEE) received the B.S. degree in Computer Science from University of Science and Technology of China in 2014 and then received the Ph.D. degree from the University of Sydney in 2019. His research interests include object detection, computer vision applications, and deep learning. His studies were published in high-quality conferences and journals like CVPR, ICCV, ECCV, IJCAI, AAAI, TIP, IJCV, and so on. He has received more then 3300 citations on the Google scholar. He also serves as a reviewer for a number of top journals like T-PAMI, TIP, TCSVT, T-CYB, etc.
\end{IEEEbiography}

\vspace{-0.5cm}
\begin{IEEEbiography}[{\includegraphics[width=1in,height=1.25in,clip,keepaspectratio]{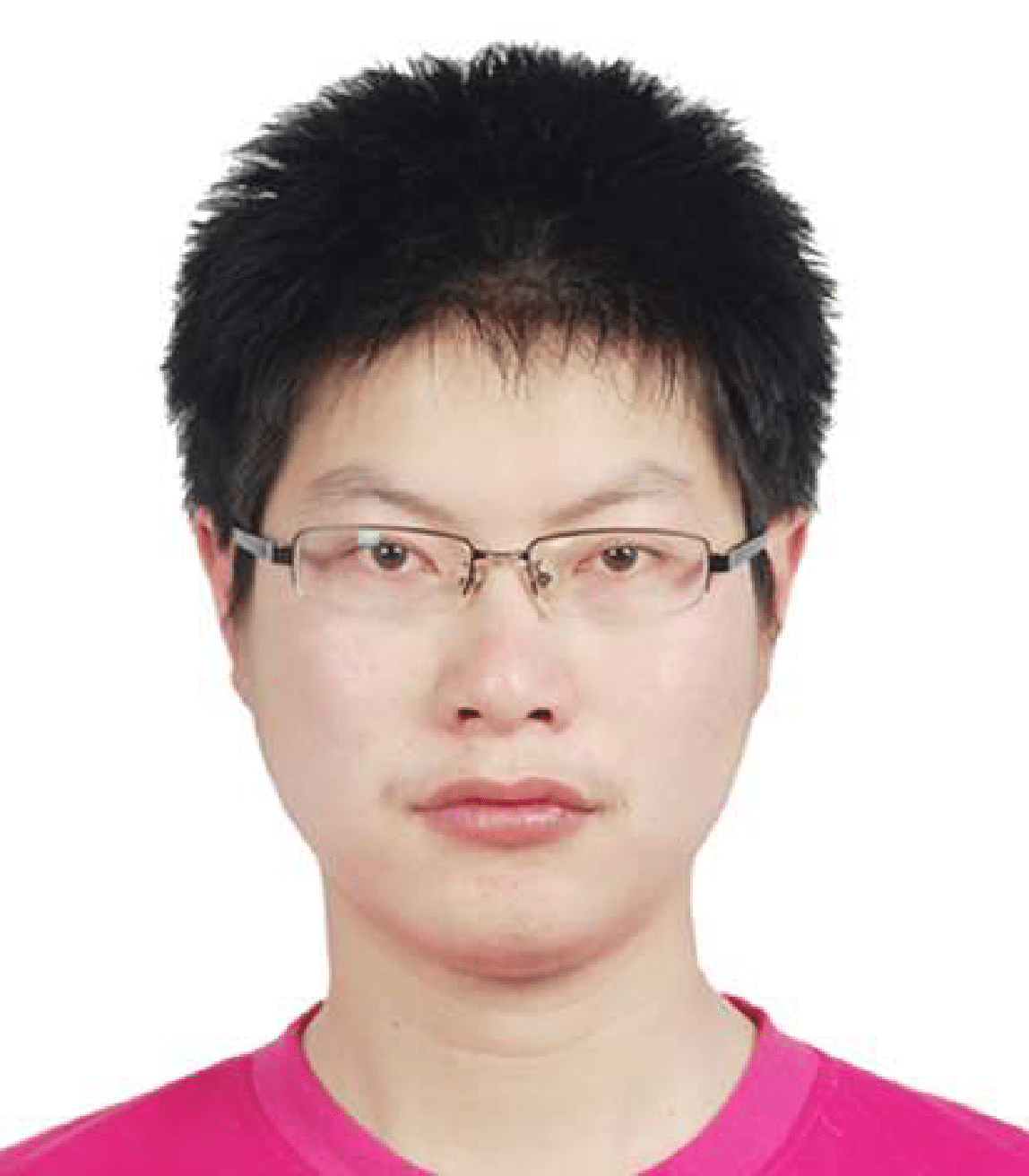}}]
{Bo Jiang} received the B.S. degrees in mathematics and applied mathematics and the  M.Eng.  and Ph.D. degree in computer science from Anhui University of China in 2009, 2012, and 2015, respectively. He is currently an associate professor in computer science at Anhui University. His current research interests include image feature extraction and matching, data representation and learning.
\end{IEEEbiography}

\vspace{-0.5cm}
\begin{IEEEbiography} [{\includegraphics[width=1in,height=1.25in,clip,keepaspectratio]{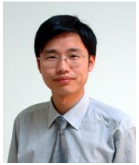}}]
{Jin Tang} received the B.Eng. Degree in automation and the Ph.D. degree in computer science from Anhui University, Hefei, China, in 1999 and 2007, respectively. He is currently a Professor with the School of Computer Science and Technology, Anhui University. His current research interests include computer vision, pattern recognition, and machine learning.
\end{IEEEbiography}

\vspace{-0.5cm}
\begin{IEEEbiography} [{\includegraphics[width=1in,height=1.25in,clip,keepaspectratio]{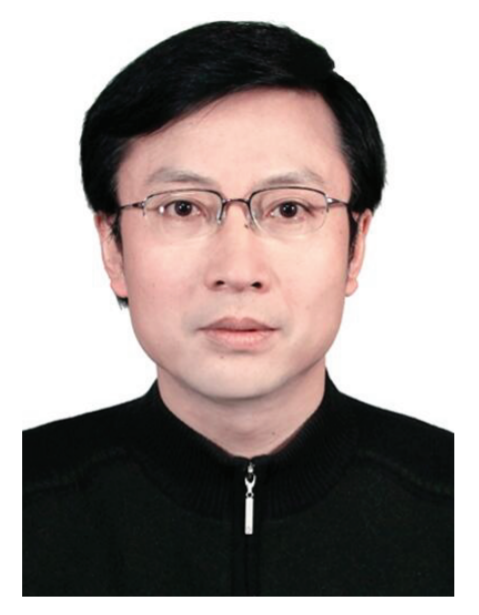}}]
{Bin Luo} (Senior Member, IEEE) received his B.Eng. degree in electronics and M.Eng. degree in computer science from Anhui University of China in 1984 and 1991, respectively. In 2002, he was awarded the Ph.D. degree in Computer Science from the University of York, the United Kingdom. He has published more than 200 papers in journal and refereed conferences. He is a professor at Anhui University of China. At present, he chairs the IEEE Hefei Subsection. He served as a peer reviewer of international academic journals such as IEEE Trans. on PAMI, Pattern Recognition, Pattern Recognition Letters, etc. His current research interests include random graph based pattern recognition, image and graph matching, spectral analysis.
\end{IEEEbiography}

\vspace{-0.5cm}
\begin{IEEEbiography}  [{\includegraphics[width=1in,height=1.25in,clip,keepaspectratio]{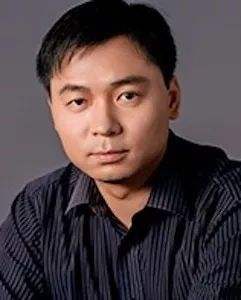}}]
{Dacheng Tao} (Fellow, IEEE) is the Inaugural Director of the JD Explore Academy and a Senior Vice President of JD.com. He is also an advisor and chief scientist of the digital sciences initiative in the University of Sydney. He mainly applies statistics and mathematics to artificial intelligence and data science, and his research is detailed in one monograph and over 200 publications in prestigious journals and proceedings at leading conferences. He received the 2015 Australian Scopus-Eureka Prize, the 2018 IEEE ICDM Research Contributions Award, and the 2021 IEEE Computer Society McCluskey Technical Achievement Award. He is a fellow of the Australian Academy of Science, AAAS, ACM and IEEE.
\end{IEEEbiography}

\end{document}